\documentclass[10pt,twocolumn,letterpaper]{article}

 \usepackage[pagenumbers]{cvpr} %

\usepackage[dvipsnames]{xcolor}

\usepackage{multirow}

\definecolor{cvprblue}{rgb}{0.21,0.49,0.74}
\usepackage[pagebackref,breaklinks,colorlinks,citecolor=cvprblue]{hyperref}

\title{SelfOcc: Self-Supervised Vision-Based 3D Occupancy Prediction}

\author{Yuanhui Huang\footnotemark[1]\quad Wenzhao Zheng\footnotemark[1] \quad Borui Zhang\quad Jie Zhou\quad Jiwen Lu\footnotemark[2] \\
Beijing National Research Center for Information Science and Technology, China \\
Department of Automation, Tsinghua University, China \\
\texttt{\{huangyh22,zhang-br21\}@mails.tsinghua.edu.cn; wenzhao.zheng@outlook.com;} \\
\texttt{\{jzhou,lujiwen\}@tsinghua.edu.cn}
}

\begin{document}

\twocolumn[{%
\renewcommand\twocolumn[1][]{#1}%
\vspace{-17mm}
\maketitle
\vspace{-12mm}
\begin{center}
    \centering
    \includegraphics[width=\linewidth]{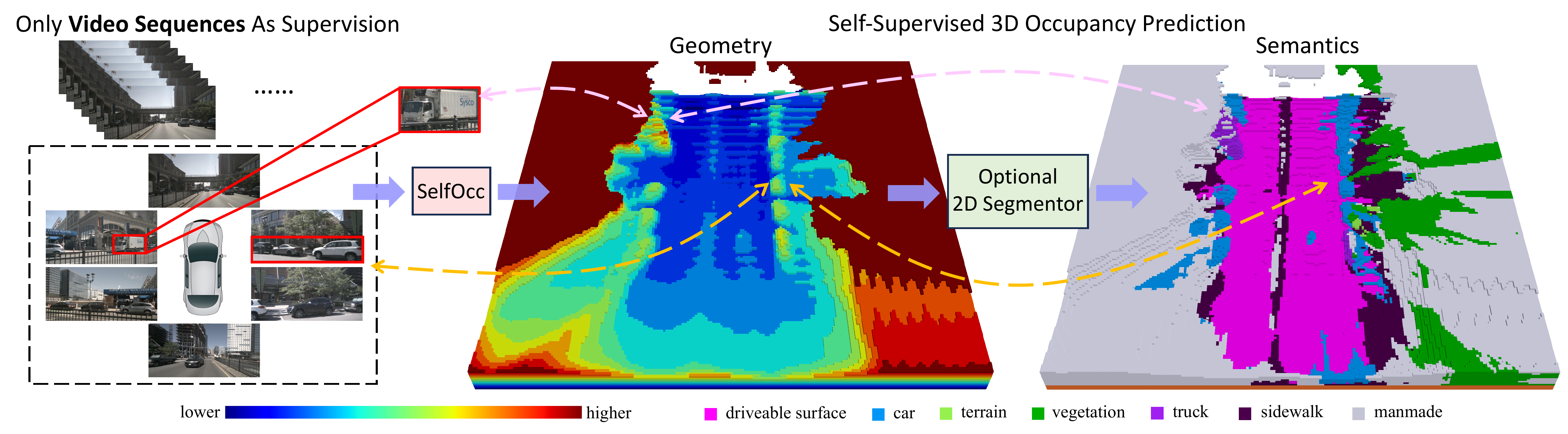}
    \vspace{-9mm}
    \captionof{figure}{
    Trained with only video sequences as supervision, our model can predict meaningful geometry for the scene given surround-camera RGB images, which can be further extended to semantic occupancy prediction if 2D segmentation maps are available e.g. from an off-the-shelf segmentor.
    This task is challenging because it completely depends on video sequences to reconstruct scenes without any 3D supervision.
    We observe that our model can produce dense and consistent occupancy prediction and even infer the back side of cars.
    }
\label{teaser}
\end{center}%
}]

\renewcommand{\thefootnote}{\fnsymbol{footnote}}
\footnotetext[1]{Equal contribution. $\dagger$Corresponding author.}
\renewcommand{\thefootnote}{\arabic{footnote}}

\begin{abstract}
3D occupancy prediction is an important task for the robustness of vision-centric autonomous driving, which aims to predict whether each point is occupied in the surrounding 3D space.
Existing methods usually require 3D occupancy labels to produce meaningful results.
However, it is very laborious to annotate the occupancy status of each voxel. 
In this paper, we propose SelfOcc to explore a self-supervised way to learn 3D occupancy using only video sequences.
We first transform the images into the 3D space (e.g., bird's eye view) to obtain 3D representation of the scene.
We directly impose constraints on the 3D representations by treating them as signed distance fields.
We can then render 2D images of previous and future frames as self-supervision signals to learn the 3D representations. 
We propose an MVS-embedded strategy to directly optimize the SDF-induced weights with multiple depth proposals.
Our SelfOcc outperforms the previous best method SceneRF by 58.7\% using a single frame as input on SemanticKITTI and is the first self-supervised work that produces reasonable 3D occupancy for surround cameras on nuScenes.
SelfOcc produces high-quality depth and achieves state-of-the-art results on novel depth synthesis, monocular depth estimation, and surround-view depth estimation on the SemanticKITTI, KITTI-2015, and nuScenes, respectively.
Code: \url{https://github.com/huang-yh/SelfOcc}.
\end{abstract}
    
\section{Introduction}
\label{sec:intro}
The vision-centric paradigm has demonstrated its potential for efficient autonomous driving~\cite{li2022bevformer,huang2023tri,hu2022uniad,jiang2023vad}, where the key procedure is to obtain a perceptual 3D representation, e.g. the bird's eye view (BEV)~\cite{ng2020bev} or tri-perspective view (TPV)~\cite{huang2023tri} of the surrounding scene given only 2D images~\cite{li2022bevformer,philion2020lss}.
Most existing methods achieve this using 3D annotations as supervision and achieve promising results on various tasks including semantic map construction~\cite{philion2020lss,ng2020bev,hu2021fiery}, 3D object detection~\cite{li2022bevformer,huang2021bevdet,li2022bevdepth,zhang2022beverse}, and 3D semantic occupancy prediction~\cite{huang2023tri,cao2022monoscene,tian2023occ3d,wang2023openoccupancy}.

Despite the strong performance of supervised 3D representation that facilitates accurate subsequent prediction and planning~\cite{hu2022uniad,tong2023scene,jiang2023vad}, it is very laborious to annotate 3D labels and thus difficult to scale to large-scale training data. 
As one workaround, data auto-labeling creates data cycles and train large networks to assist human annotation, yet it is still time-consuming and resource-hungry~\cite{qi2021offboard,yang2021auto4d,chen2022mppnet}. 
It is also noticed to be the bottleneck towards building and updating large models for autonomous driving~\cite{chen2023end}.
Therefore, it is desired to learn meaningful 3D representations only from recorded video sequences without 3D labels.
While existing works have explored learning 3D information (e.g., depth) from images~\cite{godard2019digging,cao2023scenerf,wimbauer2023behind}, they are still based on 2D representations in each monocular camera space.
It remains challenging to obtain reasonable and comprehensive 3D representations useful for further autonomous driving prediction and planning~\cite{hu2022uniad} in a self-supervised manner.

In this paper, we explore the problem of learning self-supervised 3D representation (BEV and TPV) and adopt 3D occupancy prediction as the main task to evaluate the quality of the learned 3D representation, which aims to classify each point in the 3D space into occupied or empty. 
We first lift 2D image features into 3D space with deformable attention layers~\cite{li2022bevformer,huang2023tri} to enable feature interactions in the 3D space and avoid ambiguities from multiple cameras.
Inspired by recent advances in neural implicit surface reconstruction~\cite{wang2021neus,oechsle2021unisurf}, we then transform the 3D representation into a signed distance function (SDF) field to allow for meaningful regularization and straightforward determination of occupancy boundaries.
When exploiting the temporal consistency inherent in video sequences to optimize the SDF field, we propose an MVS-embedded strategy that directly optimizes the SDF-induced weight values with multiple depth proposals.
Combined with the temporal supervision scheme and loss formulation tailored for 3D occupancy prediction, our method can reconstruct meaningful 3D occupancy in a self-supervised manner.

We conduct extensive experiments on various datasets to demonstrate the effectiveness of the proposed SelfOcc.
We apply SelfOcc to two types of 3D representations BEV and TPV.
For surround-view 3D occupancy prediction, SelfOcc is the first self-supervised method that is able to produce reasonable occupancy results using only video supervision (with an IoU of 45.01 and mIoU of 9.30 on Occ3D~\cite{tian2023occ3d}).
For monocular 3D occupancy prediction, SelfOcc outperforms the previous best method SceneRF~\cite{cao2023scenerf} by 58.7\% with an IoU of 21.97 over 13.84 on SemanticKITTI~\cite{behley2019semantickitti}.
To further evaluate the usefulness of the learned 3D representations, we render depth maps for each camera from the self-supervised BEV and TPV representation.
Our SelfOcc outperforms the mainstream method SceneRF~\cite{cao2023scenerf}, MonoDepth2~\cite{godard2019digging}, and SurroundDepth~\cite{wei2023surrounddepth} on novel depth synthesis, monocular depth estimation, and surround-view depth estimation on the SemanticKITTI~\cite{behley2019semantickitti}, KITTI-2015~\cite{kitti-2015}, and nuScenes~\cite{caesar2020nuscenes}, respectively.

\section{Related Work}
\label{sec: related work}

\textbf{3D Occupancy Prediction:} 
While detection, tracking, prediction and planning have long been the focal tasks and composed a standard pipeline for autonomous driving, 3D occupancy prediction~\cite{huang2023tri} or semantic scene completion~\cite{song2017semantic} has recently gained attention as a more fundamental task encompassing the most fine-grained prediction of both occupancy and semantics of the environment.
Pioneering works rely on 3D inputs such as  depth~\cite{SATNet,Li2019ddr,TS3D,Li2020AttentionbasedMF,PALNet,aicnet,Cheng2020S3CNetAS}, occupancy grids~\cite{TS3D,lmscnet,Wu2020SCFusionRI,yan2021JS3CNet}, point cloud~\cite{Zhong2020SemanticPC,localDif}, or truncated signed distance function (TSDF)~\cite{song2017semantic,zhang2018efficient,EdgeNet,dourado360degreeSSC,Chen20193DSS,ForkNet,CCPNet,PALNet,3DSketch,Cheng2020S3CNetAS,SISNet}.
Cao et al.~\cite{cao2022monoscene} and Huang et al.~\cite{huang2023tri} first attempt to predict semantic occupancy from image input only.
Recent advancements related to 3D occupancy prediction include modality fusion~\cite{wang2023openoccupancy}, multi-task learning~\cite{tong2023scene}, and end-to-end autonomous driving~\cite{hu2022uniad}.
Despite the promising performance, all these methods require 3D ground truth from laborious annotation for supervision.
Most related to our work is BTS~\cite{wimbauer2023behind} and SceneRF~\cite{cao2023scenerf}, which learn 3D occupancy in a self-supervised manner in monocular scenarios.
However, it is nontrivial to adapt them to surround views, while our work can inherently handle monocular and surround cases.

\textbf{Neural Radiance Fields:}
One of the most predominant paradigms for self-supervised 3D reconstruction is neural radiance fields~\cite{mildenhall2021nerf}, which learn a multi-layer perceptron (MLP) per scene to map spatial locations to radiance and density values.
To synthesize a novel view, volume rendering~\cite{max1995optical} is often leveraged which queries the MLP at multiple locations along rays and integrates radiance over density-induced weights.
Training of NeRFs~\cite{mildenhall2021nerf,barron2021mip,barron2022mip,chen2022tensorf,muller2022instant,fridovich2022plenoxels} depends on a large number of images of the same scene with ground truth poses.
To facilitate sparse view reconstruction and generalization for object-level data, several methods~\cite{pixelnerf,muller2022autorf,visionnerf} propose to condition NeRFs on image features which establish the foundation for online NeRFs.
Recently, this boundary has been further extended to scene-level online reconstruction by \cite{mine2021,wimbauer2023behind,cao2023scenerf}.
However, these works still use image features as conditions for NeRFs, ignoring 3D feature encoding crucial for 3D perception, while our method learns 3D representations explicitly. 

\textbf{Self-supervised Depth Prediction:}
Training of NeRFs is time-consuming and prone to overfitting due to geometry-appearance coupling, which can be alleviated with additional depth supervision from separately trained depth prediction modules~\cite{dsnerf,roessle2022depthpriorsnerf,wei2021nerfingmvs} or LiDAR data~\cite{urbannerf}.
To integrate explicit depth optimization in self-supervised generalizable NeRFs, BTS~\cite{wimbauer2023behind} and SceneRF~\cite{cao2023scenerf} leverage the reprojection loss from self-supervised depth estimation literature~\cite{zhou2017unsupervised,godard2017unsupervised,godard2019digging,zhan2018unsupervised}, which minimizes the discrepancy between the target pixel and its warped counterpart.
However, the extra degrees of freedom from volume rendering and the local receptive field from bilinear interpolation may still hinder depth optimization, while we propose a novel MVS-embedded approach to learn depth in NeRFs.
\section{Proposed Approach}
\label{sec: method}

In this section, we present a self-supervised vision-based 3D occupancy prediction method for autonomous driving.

\subsection{From Image to Occupancy}
\label{method sub: 1}
To understand the complicated interactions among traffic participants and their surrounding environment, it is fundamental for autonomous driving systems to reconstruct the accurate 3D structure of the scene.
3D occupancy prediction is one prevalent proxy for scene reconstruction given its fine granularity and comprehensiveness, which aims at producing a voxelized prediction $\mathbf{O}\in \mathcal{C}^{H\times W\times D}$ encoding per-voxel occupancy (and semantic) information.
Here $H, W, D$ and $\mathcal{C}$ denote the resolution of the occupancy grid and the set of predefined classes.
First proposed by MonoScene~\cite{cao2022monoscene}, vision-based methods have become a promising approach to 3D occupancy prediction because of their low cost and effectiveness compared with 3D-based methods. %
Vision-based methods generally learn a mapping $\mathcal{M}=\mathcal{D}\circ\mathcal{F}\circ\mathcal{E}$ from RGB images $\mathbf{I} = \{\mathbf{I}_{n} | n = 1, ..., N\}$ captured by $N$ cameras to 3D occupancy $\mathbf{O}$:
\begin{equation}\label{eq: overall}
    \mathbf{O} = \mathcal{M}(\mathbf{I}) = \mathcal{D}(\mathcal{F}(\mathcal{E}(\mathbf{I}_1), ..., \mathcal{E}(\mathbf{I}_N))),
\end{equation}
where a 2D backbone $\mathcal{E}$ first encodes $N$ input images into multi-scale image features $\mathbf{F} = \mathcal{E}(\mathbf{I})$, and a 3D encoder $\mathcal{F}$ further lifts 2D features into 3D representation $\mathbf{R} = \mathcal{F}(\mathbf{F})$ through attention mechanism~\cite{li2022bevformer,huang2023tri}, lift-splat-shoot (LLS)~\cite{li2022bevdepth} or identical transformation for monocular scenario~\cite{cao2022monoscene}, followed by a decoder network $\mathcal{D}$ to generate the final occupancy prediction result $\mathbf{O} = \mathcal{D}(\mathbf{R})$.

\begin{figure}[t]
\centering
\includegraphics[width=0.475\textwidth]{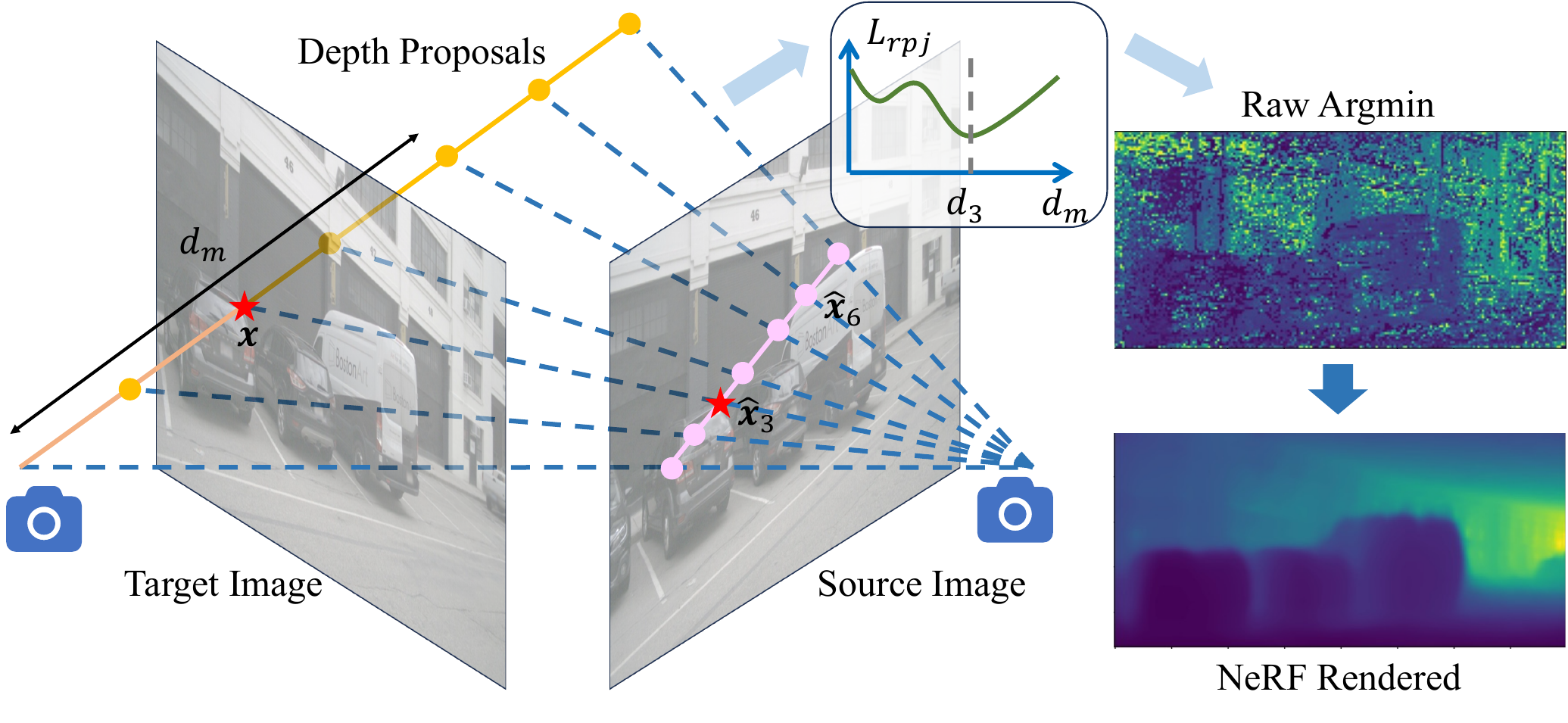}
\vspace{-9mm}
\caption{
Our MVS-embedded strategy effectively enlarges the receptive field of the depth optimization process across the whole epipolar line, which provides a good depth prior (Raw Argmin).
}
\label{fig:comparisons}
\vspace{-5mm}
\end{figure}

Although recent supervised vision-based approaches have achieved encouraging performance on 3D occupancy prediction given only visual inputs at inference, they still rely on 3D (semantic) supervision during training, e.g. LiDAR point cloud or dense occupancy annotation. %
On the contrary, self-supervised counterparts~\cite{cao2023scenerf,wimbauer2023behind} take only temporal correspondences inherent in video sequences as supervision to predict meaningful occupancy, and therefore can effortlessly exploit the vast amount of unlabeled driving images.
However, existing self-supervised methods~\cite{cao2023scenerf,wimbauer2023behind} consider only one monocular camera and decode forward occupancy $\mathbf{O}_{front}$ directly from 2D image features by instantiating the 3D encoder $\mathcal{F}$ with identical transformation: %
\begin{equation}\label{eq: mono recon}
    \mathbf{O}_{front} = \mathcal{D}(\mathcal{F}(\mathcal{E}(\mathbf{I}_{mono}))) = \mathcal{D}(\mathcal{E}(\mathbf{I}_{mono})).
\end{equation}
Despite being straightforward, ignoring $\mathcal{F}$ may cause multi-camera inconsistency in the surround view scenario and inadequate geometry reasoning in the 3D space.

To this end, we propose to transform the images into the BEV or TPV space to obtain a 3D representation of the scene in order to enable 3D feature interactions even in the monocular setting and also avoid ambiguities from multiple cameras.
Similar to the supervised literature~\cite{li2022bevformer,huang2023tri}, we leverage the deformable cross-attention ($\rm CA$) to adaptively aggregate information from the image features $\mathbf{F}$, in which a set of learnable 3D tokens $\mathbf{Q}$ serve as queries and the corresponding local image features serve as keys and values.
Each of these learnable tokens represents a pillar area as in BEV or TPV representation.
Moreover, we determine the correspondences between 3D tokens and image features with the projection matrices $\mathbf{T} = \{\mathbf{T}_n | n = 1, ..., N\}$ from the ego car to pixel coordinates. %
We interleave deformable self-attention ($\rm SA$), deformable cross-attention ($\rm CA$) and feed forward network ($\rm FFN$) to build the 3D encoder $\mathcal{F}$.
\begin{equation}\label{eq: block of F}
    \mathcal{F}_l = {\rm FFN}_l({\rm CA}_l({\rm SA}_l(\mathbf{Q}_l); \mathbf{F}, \mathbf{T})),
\end{equation}
where the subscription $l$ represents the $l$-th block.

\begin{figure*}[t]
\centering
\includegraphics[width=\textwidth]{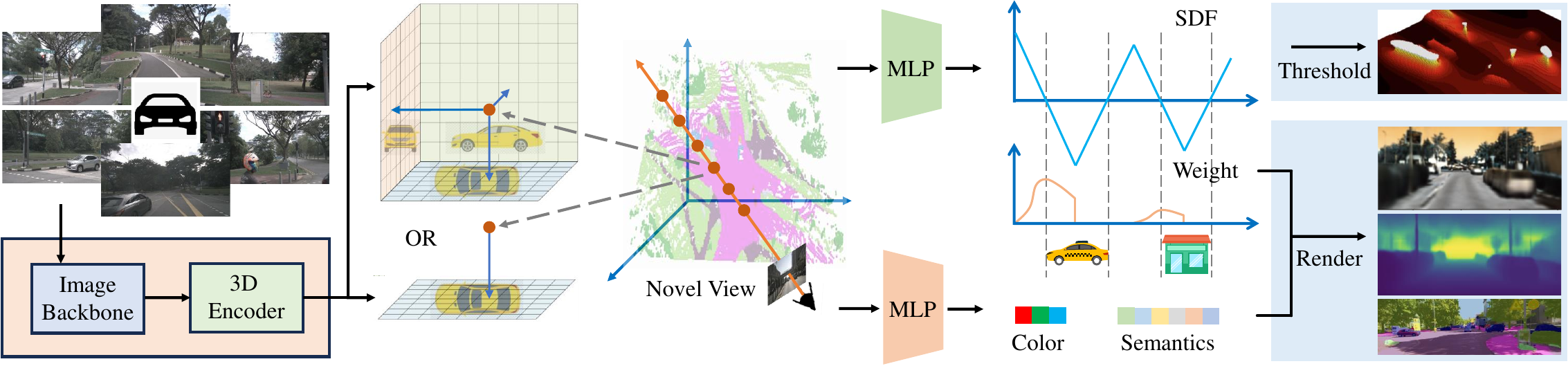}
\vspace{-7mm}
\caption{
Framework of the proposed SelfOcc for self-supervised vision-based 3D occupancy prediction.
We employ an image backbone and a 3D encoder to generate 3D representations as in BEVFormer~\cite{li2022bevformer} or TPVFormer~\cite{huang2023tri}.
To render a novel view, we apply a lightweight MLP on the 3D features to predict the SDF values, color and semantic vectors.
We then perform volume rendering to synthesize color, depth and semantic views.
We use simple 0-thresholding to predict the occupancy volume.
}
\label{fig:framework}
\vspace{-5mm}
\end{figure*}

\subsection{From Occupancy to Image}
\label{method sub: 2}
Because 3D ground truth is unavailable, we project occupancy $\mathbf{O}$ from Eq.~(\ref{eq: overall}) back into 2D views and leverage multi-view consistency inherent in video sequences to optimize a self-supervised vision-based occupancy prediction system.
Similar to previous work~\cite{cao2023scenerf,wimbauer2023behind}, we employ differentiable volume rendering~\cite{mildenhall2021nerf} to synthesize color and depth views, which enables the seamless integration of gradient information into the rendering pipeline and effective exploitation of supervision from multiple viewpoints.%

We first transform the 3D representation $\mathbf{R}$ into an SDF field $\mathbf{S}\in \mathbb{R}^{H\times W\times D}$, which indicates the distance of every voxel center to the nearest surface of an object, by implementing the decoder network $\mathcal{D}$ with an MLP.
We use different MLPs for different heights for BEV representation, while we construct a 3D feature volume from TPV representation with broadcasting and summation~\cite{huang2023tri} before feeding the feature volume through a single MLP.
For a continuous 3D coordinate $\mathbf{p}$, we employ bicubic interpolation ($\rm BI$) to predict the SDF value $s_{\mathbf{p}}$ and determine its occupancy status $o_{\mathbf{p}}$ according to the sign of $s_{\mathbf{p}}$: %
\begin{equation} \label{eq: interp sdf}
    o_{\mathbf{p}} = {\rm sgn}(s_{\mathbf{p}}),\quad s_{\mathbf{p}} = {\rm BI}(\mathbf{S}, \mathbf{p}) = {\rm BI}(\mathcal{D}(\mathbf{R}), \mathbf{p}).
\end{equation}
The preference for SDF field over density field as in \cite{cao2023scenerf,wimbauer2023behind} is based on two considerations: 1) An SDF field has more explicit physical meaning compared with a density field and inherently holds assumptions about the gradient magnitudes~\cite{gropp2020implicit}, thus allowing for easier regularization and optimization. 2) The signed nature of SDF facilitates straightforward determination of whether a point lies inside or outside a surface, thus enabling accurate differentiation of intricate geometries. 
Similar to SDF values, we can also decode color $\mathbf{c}$ from $\mathbf{R}$ by using a separate MLP $\mathcal{D}_c$ as the decoder: %
\begin{equation}\label{eq: interp color}
    \mathbf{c}_{\mathbf{p}} = {\rm BI}(\mathbf{C}, \mathbf{p}) = {\rm BI}(\mathcal{D}_c(\mathbf{R}), \mathbf{p}).
\end{equation}

Take the volume rendering process of one single ray as an example.
We first determine the origin and direction of the ray and uniformly sample $M$ points $\mathbf{P} = \{\mathbf{p}_m | m=1,..., M\}$ between the origin and the intersection of the ray with the border of the 3D representation.
The SDF values of these points $\mathbf{s} = \{s_m | m=1,...,M\}$ are then calculated with Eq.~(\ref{eq: interp sdf}).
According to NeuS~\cite{wang2021neus}, the discrete opacity value $\alpha_m$, which is the probability of the ray terminating between $\mathbf{p}_m$ and $\mathbf{p}_{m+1}$, can be derived with
\begin{equation}
    \alpha_m = {\rm max}(\frac{\Phi_a(s_m) - \Phi_a(s_{m+1})}{\Phi_a(s_m)}, 0),
\end{equation}
where $\Phi_a(x) = (1+e^{-ax})^{-1}$ is the sigmoid function with $a$ being a learnable parameter.
Further, the probability of the ray passing through $\mathbf{p}_m$, i.e. the discrete accumulated transmittance, can be defined by $T_m = \prod_{i=1}^{m-1}(1 - \alpha_i)$.
Finally, we can calculate the probability of the ray ending at $\mathbf{p}_m$ after emission as $w_m = T_m \alpha_m$, and the rendering process is an integration of per-point attribute over the probability distribution $\mathbf{w}=\{w_m | m = 1, ..., M\}$.
Eq.~(\ref{eq: color depth int}) calculates the rendered color $\mathbf{c}_r$ and depth $d_r$ of the ray:
\begin{equation}\label{eq: color depth int}
    \mathbf{c}_r = \sum_{m=1}^{M} w_m \mathbf{c}_m, \quad d_r = \sum_{m=1}^{M} w_m d_m,
\end{equation}
where $\mathbf{c}_m$, $d_m$ denote the color and depth of the $m$-th point.
Figure~\ref{fig:framework} shows the overall framework of our method.

\subsection{Occupancy-Oriented Supervision}
\label{method sub: 3}
In this section, we elaborate on the supervision formulation that facilitates self-supervised vision-based 3D occupancy prediction for autonomous driving.

\textbf{MVS-embedded Depth Learning.}
Depth supervision enhances the convergence of neural radiance fields~\cite{wei2021nerfingmvs}, enabling faster and more effective learning of accurate geometry, especially with sparse views.
Self-supervised depth estimation in NeRFs usually relies on the photometric reprojection loss $L_{rpj}$ which aims to maximize the similarity between the target image $\mathbf{I}_t$ and sampled pixels on the source image $\mathbf{I}_s$ by refining the depth prediction $\mathbf{D}(\mathbf{I}_t;\boldsymbol{\theta})$:
\begin{equation}\label{eq: reproj}
\begin{aligned}
    L_{rpj}(\mathbf{x}, \mathbf{I}_t, \mathbf{I}_s; &\boldsymbol{\theta}) = \Vert\mathbf{I}_t(\mathbf{x}) - \mathbf{I}_s(\hat{\mathbf{x}})\Vert \\ &= \Vert\mathbf{I}_t(\mathbf{x}) - \mathbf{I}_s(\pi(\mathbf{x}, \mathbf{D}(\mathbf{I}_t;\boldsymbol{\theta})|_{\mathbf{x}}, \boldsymbol{\Pi}))\Vert,
\end{aligned}
\end{equation}
where $\boldsymbol{\theta}$, $\mathbf{x}$, $\hat{\mathbf{x}}$, $\boldsymbol{\Pi}$ denote the learnable parameters, a random pixel on the target image, the warped pixel on the source image and the projection matrix from the target to the source image coordinate, respectively.
$\mathbf{I}(\mathbf{x})$ and $\mathbf{D}|_{\mathbf{x}}$ denote bilinear interpolation of corresponding 2D tensors, and $\pi(\mathbf{x}, d_\mathbf{x}, \boldsymbol{\Pi})$ warps a pixel $\mathbf{x}$ to another image according to the depth $d_\mathbf{x}$ and projection matrix $\boldsymbol{\Pi}$.
The inherent local receptive field of the bilinear interpolation $\mathbf{I}_s(\hat{\mathbf{x}})$ has an adverse effect on the optimization property of $L_{rpj}$ since the differences between $\mathbf{I}_t(\mathbf{x})$ and the colors of the four adjacent grid pixels of $\hat{\mathbf{x}}$ are the only reference information for the optimization of $\boldsymbol{\theta}$, which could be easily misled by poor initialization or low-texture regions.

\definecolor{nbarrier}{RGB}{255, 120, 50}
\definecolor{nbicycle}{RGB}{255, 192, 203}
\definecolor{nbus}{RGB}{255, 255, 0}
\definecolor{ncar}{RGB}{0, 150, 245}
\definecolor{nconstruct}{RGB}{0, 255, 255}
\definecolor{nmotor}{RGB}{200, 180, 0}
\definecolor{npedestrian}{RGB}{255, 0, 0}
\definecolor{ntraffic}{RGB}{255, 240, 150}
\definecolor{ntrailer}{RGB}{135, 60, 0}
\definecolor{ntruck}{RGB}{160, 32, 240}
\definecolor{ndriveable}{RGB}{255, 0, 255}
\definecolor{nother}{RGB}{139, 137, 137}
\definecolor{nsidewalk}{RGB}{75, 0, 75}
\definecolor{nterrain}{RGB}{150, 240, 80}
\definecolor{nmanmade}{RGB}{213, 213, 213}
\definecolor{nvegetation}{RGB}{0, 175, 0}
\definecolor{nothers}{RGB}{0, 0, 0}

\begin{table*}[t]
\footnotesize
\setlength{\tabcolsep}{0.004\linewidth}
\centering
\caption{
\textbf{3D occupancy prediction performance on the Occ3D-nuScenes~\cite{tian2023occ3d} dataset.} 
Cons. veh. and drive. surf. represent construction vehicle and driveable surface, respectively.
Our method learns meaningful geometry and reasonable semantics compared with 3D-supervised methods, and achieves even higher IoU than LiDAR-supervised TPVFormer~\cite{huang2023tri}.
} 
\vspace{-3mm}
\begin{tabular}{l | c c c | c c c c c c c c c c c c c c c c c}

    \toprule
    Method 
    & \rotatebox{90}{Supervision}
    & \rotatebox{90}{IoU}
    & \rotatebox{90}{mIoU}
    & \rotatebox{90}{\textcolor{nothers}{$\blacksquare$} others} 
    & \rotatebox{90}{\textcolor{nbarrier}{$\blacksquare$} barrier} %
    & \rotatebox{90}{\textcolor{nbicycle}{$\blacksquare$} bicycle} %
    & \rotatebox{90}{\textcolor{nbus}{$\blacksquare$} bus} %
    & \rotatebox{90}{\textcolor{ncar}{$\blacksquare$} car} %
    & \rotatebox{90}{\textcolor{nconstruct}{$\blacksquare$} cons. veh.} %
    & \rotatebox{90}{\textcolor{nmotor}{$\blacksquare$} motorcycle} %
    & \rotatebox{90}{\textcolor{npedestrian}{$\blacksquare$} pedestrian} %
    & \rotatebox{90}{\textcolor{ntraffic}{$\blacksquare$} traffic cone} %
    & \rotatebox{90}{\textcolor{ntrailer}{$\blacksquare$} trailer} %
    & \rotatebox{90}{\textcolor{ntruck}{$\blacksquare$} truck} %
    & \rotatebox{90}{\textcolor{ndriveable}{$\blacksquare$} drive. surf.} %
    & \rotatebox{90}{\textcolor{nother}{$\blacksquare$} other flat} %
    & \rotatebox{90}{\textcolor{nsidewalk}{$\blacksquare$} sidewalk} %
    & \rotatebox{90}{\textcolor{nterrain}{$\blacksquare$} terrain} %
    & \rotatebox{90}{\textcolor{nmanmade}{$\blacksquare$} manmade} %
    & \rotatebox{90}{\textcolor{nvegetation}{$\blacksquare$} vegetation} \\ %
    \midrule
    
    MonoScene~\cite{cao2022monoscene} & 3D & - & 6.06  & 1.75 & 7.23 & 4.26 & 4.93 & 9.38 & 5.67 & 3.98 & 3.01 & 5.90 & 4.45 & 7.17 & 14.91 & 6.32 & 7.92 & 7.43 & 1.01 & 7.65 \\
    OccFormer~\cite{zhang2023occformer} & 3D & - & 21.93 & 5.94 & 30.29 & 12.32 & 34.40 & 39.17 & 14.44 & 16.45 & 17.22 & 9.27 & 13.90 & 26.36 & 50.99 & 30.96 & 34.66 & 22.73 & 6.76 & 6.97 \\
    BEVFormer~\cite{li2022bevformer} & 3D & - & 26.88 & 5.85 & 37.83 & \underline{17.87} & \underline{40.44} & \underline{42.43} & 7.36 & \underline{23.88} & \underline{21.81} & \underline{20.98} & 22.38 & 30.70 & {55.35} & 28.36 & 36.0 & 28.06 & \underline{20.04} & \bf{17.69} \\
    CTF-Occ~\cite{tian2023occ3d} & 3D & - & \bf{28.53} & \bf{8.09} & \bf{39.33} & \bf{20.56} & 38.29 & 42.24 & \underline{16.93} & \bf{24.52} & \bf{22.72} & \bf{21.05} & \underline{22.98} & \underline{31.11} & 53.33 & \underline{33.84} & \bf{37.98} & \bf{33.23} & \bf{20.79} & 18.0 \\
    TPVFormer~\cite{huang2023tri} & 3D & -  & \underline{27.83} & \underline{7.22} & \underline{38.90} & 13.67 & \bf{40.78} & \bf{45.90} & \bf{17.23} & 19.99 & 18.85 & 14.30 & \bf{26.69} & \bf{34.17} & \bf{55.65} & \bf{35.47} & \underline{37.55} & \underline{30.70} & 19.40 & 16.78 \\
    TPVFormer~\cite{huang2023tri} & L & 17.20 & 13.57 & 0.00 & 14.80 & 9.36 & 21.27 & 16.81 & 14.45 & 13.76 & 11.23 & 5.32 & 16.05 & 19.73 & 10.75 & 9.43 & 9.50 & 11.16 & 16.51 & \underline{17.04} \\
    \bf{SelfOcc (BEV)} & C & \underline{44.33} & 6.76 & 0.00 & 0.00  & 0.00  & 0.00  & 9.82  & 0.00  & 0.00 & 0.00  & 0.00  & 0.00  & 6.97  & 47.03  & 0.00  & 18.75  & 16.58  & 11.93  & 3.81  \\
    \bf{SelfOcc (TPV)} & C & \bf{45.01} & 9.30 & 0.00 & 0.15 & 0.66 & 5.46 & 12.54 & 0.00 & 0.80 & 2.10 & 0.00 & 0.00 & 8.25 & \underline{55.49} & 0.00 & 26.30 & 26.54 & 14.22 & 5.60 \\
\bottomrule
\end{tabular}
\vspace{-5mm}
\label{table:occ3d-nus}
\end{table*}

Therefore, we propose an MVS-embedded strategy to extend the receptive field of the depth optimization for a ray across the whole epipolar line.
Specifically, we first take the uniformly spaced depths $\mathbf{d}=\{d_m | m = 1, ..., M\}$ associated with $\mathbf{P} = \{\mathbf{p}_m | m = 1, ..., M\}$ as depth proposals.
Then we define $L_{mvs}$ as an average of the dissimilarity at each proposal $d_m$ with weight $w_m$ from volume rendering:
\begin{equation}\label{eq: mvs depth}
\begin{aligned}
    L_{mvs}&(\mathbf{x},\mathbf{I}_t,\mathbf{I}_s;\boldsymbol{\theta}) \\ &= \sum_{m=1}^{M}w_m(\boldsymbol{\theta}) \Vert\mathbf{I}_t(\mathbf{x}) - \mathbf{I}_x(\pi(\mathbf{x}, d_m, \boldsymbol{\Pi}))\Vert.
\end{aligned}
\end{equation}
As Figure~\ref{fig:comparisons} shows, $L_{mvs}$ enlarges the receptive field of depth optimization by associating $\boldsymbol{\theta}$ with multiple depth proposals along the ray, thus stabilizing and facilitating self-supervised depth optimization in NeRFs. 
Similar to Monodepth2~\cite{godard2019digging}, we formulate the final depth loss by taking the minimum of two $L_{mvs}$s using the previous image $\mathbf{I}_{t-1}$ or the next image $\mathbf{I}_{t+1}$ as the source image and leverage the auto-masking strategy to filter out low texture regions:
\begin{equation}\label{eq: loss depth}
\begin{aligned}
    L_{dep}(\mathbf{x},\mathbf{I}_t) = {\rm automask}({\rm min}(L_{mvs}^{(t-1)}, L_{mvs}^{(t+1)})),
\end{aligned}
\end{equation}
where we omit the learnable parameters $\boldsymbol{\theta}$. %

\textbf{Supervision from temporal frames.}
Unlike depth estimation which predicts only the visible surfaces, 3D occupancy prediction reconstructs the complete geometry of the scene which requires self-supervision signals from multiple viewpoints as in NeRFs.
Considering the limited overlapping field of view for cameras, we take advantage of the temporal frames as supervision.
Given a video sequence $\mathbf{V}=\{\mathbf{V}_1, ..., \mathbf{V}_T\}$ with $T$ frames and the current timestamp $t$, we use $\mathbf{V}_t$ as the input to our model $\mathcal{M}$ to predict the SDF and radiance fields $\{\mathbf{S}_t, \mathbf{C}_t\}$. 
Then we decide whether to use a temporal frame as supervision with probability $p$.
If positive, we randomly sample one frame $t'$ for supervision satisfying that the distance between the ego cars at timestamps $t$ and $t'$ falls inside $[l_1, l_2]$, which guarantees diversity and adheres to the range of the 3D representation.
Otherwise, we use the current frame $t$ for supervision.

\textbf{SDF Field Regularization.}
Neural radiance fields with sparse views suffer from overfitting because of a large solution set~\cite{niemeyer2022regnerf}, which is exacerbated in autonomous driving with outward-facing cameras.
Therefore, it is crucial to regularize the SDF field in order to restrict the solution set.
To learn a smooth SDF field, we minimize the magnitude of the second-order derivatives of the SDF field with a Hessian loss~\cite{zhang2022critical} $L_H(\mathbf{p})=\Vert{\rm \mathbf{H}}(\mathbf{p})\Vert_1$, where $\Vert\cdot\Vert_1$ is the element-wise matrix 1-norm, and ${\rm \mathbf{H}}(\mathbf{p})$ denotes the Hessian matrix of the SDF field with respect to the 3D coordinates at location $\mathbf{p}$.
To impose sparsity of occupancy in the invisible area such as those blocked by a wall, we propose a simple yet effective regularization $L_s(\mathbf{p}) = {\rm max}(-s_{\mathbf{p}}, 0)$, which encourages negative SDF values to become larger.
In addition, we include the Eikonal term~\cite{gropp2020implicit} $L_E(\mathbf{p}) = |\Vert\nabla {\rm s}(\mathbf{p})\Vert - 1|$ for compliance with physical meaning of SDF.
For optional semantic prediction, we use 2D segmentation maps from off-the-shelf segmentors as supervision similar to colors.

Til now, we define the training loss $L$ as follows:
\begin{equation}\label{eq: tot loss}
    L = L_{dep} + \lambda_c L_{rgb} + \lambda_e L_E + \lambda_H L_H + \lambda_s L_s
\end{equation}
where the four $\lambda$s are hyperparameters and $L_{rgb}$ measures the dissimilarity between the rendered color $\mathbf{c}_r$ of a ray and the ground truth color on the target image.
For semantic prediction, we optionally add a semantic loss $L_{sem}$ similar to $L_{rbg}$ and an off-the-shelf 2D segmentor.
\definecolor{nbarrier}{RGB}{255, 120, 50}
\definecolor{nbicycle}{RGB}{255, 192, 203}
\definecolor{nbus}{RGB}{255, 255, 0}
\definecolor{ncar}{RGB}{0, 150, 245}
\definecolor{nconstruct}{RGB}{0, 255, 255}
\definecolor{nmotor}{RGB}{200, 180, 0}
\definecolor{npedestrian}{RGB}{255, 0, 0}
\definecolor{ntraffic}{RGB}{255, 240, 150}
\definecolor{ntrailer}{RGB}{135, 60, 0}
\definecolor{ntruck}{RGB}{160, 32, 240}
\definecolor{ndriveable}{RGB}{255, 0, 255}
\definecolor{nother}{RGB}{139, 137, 137}
\definecolor{nsidewalk}{RGB}{75, 0, 75}
\definecolor{nterrain}{RGB}{150, 240, 80}
\definecolor{nmanmade}{RGB}{213, 213, 213}
\definecolor{nvegetation}{RGB}{0, 175, 0}
\definecolor{nothers}{RGB}{0, 0, 0}

\section{Experiments}

\begin{table}
    \small
	\centering
	\caption{\textbf{3D occupancy prediction performance on the SemanticKITTI~\cite{behley2019semantickitti} dataset.} MonoScene* is supervised by depth predictions of Monodepth2~\cite{godard2019digging} trained with ground-truth poses. %
	}
	\vspace{-3mm}
	\setlength{\tabcolsep}{0.015\linewidth}
	{
		\begin{tabular}{l|c c c|ccc}
			\toprule
			\multirow{2}[1]{*}{Method} & \multicolumn{3}{c|}{Supervision} & \multirow{2}[1]{*}{IoU} & \multirow{2}[1]{*}{Prec.} & \multirow{2}[1]{*}{Rec.} \\
			&  3D & Depth & Image &  &  &  \\
			\midrule
			MonoScene~\cite{cao2022monoscene} & \checkmark & & & \bf{37.14} & \bf{49.90} & 59.24 \\ 
			\midrule 
			LMSCNet$^{\text{rgb}}$~\cite{lmscnet} & & \checkmark & & 12.08 & 13.00 & \underline{63.16} \\ 
			3DSketch$^{\text{rgb}}$~\cite{3DSketch} & & \checkmark &  & 12.01 & 12.95 & 62.31 \\
			AICNet$^{\text{rgb}}$~\cite{aicnet}  & & \checkmark & & 11.28 & 11.84 & \bf{70.89} \\ 
			MonoScene~\cite{cao2022monoscene} &  & \checkmark & & 13.53 & 16.98 & 40.06 \\ 
			\midrule 
			
			MonoScene*~\cite{cao2022monoscene}  & &  & \checkmark & 11.18 &  13.15 & 40.22 \\
			SceneRF~\cite{cao2023scenerf} & & & \checkmark & 13.84 & 17.28 & 40.96 \\
            \bf{SelfOcc (BEV)} & & & \checkmark & 20.95 & \underline{37.27} & 32.37 \\
            \bf{SelfOcc (TPV)} & & & \checkmark & \underline{21.97} & 34.83 & 37.31 \\
			\bottomrule
		\end{tabular}
	}
\label{tab:semantickitti occ}
\vspace{-5mm}
\end{table}

\begin{table*}[t]
	\centering
	\caption{\textbf{Novel depth synthesis on SemanticKITTI~\cite{behley2019semantickitti} and nuScenes~\cite{caesar2020nuscenes}.} We use the results reported in SceneRF~\cite{cao2023scenerf} and all methods are trained using only images with ground truth poses. We adapt SceneRF~\cite{cao2023scenerf} to the nuScenes dataset based on the official implementation.}
  \vspace{-3mm}
	\setlength{\tabcolsep}{0.013\linewidth}
	\resizebox{0.95\textwidth}{!}{
		\begin{tabular}{l|c|ccccccc}
			\toprule
			Method & Dataset & Abs Rel$\downarrow$ & Sq Rel$\downarrow$ & RMSE$\downarrow$ & RMSE log$\downarrow$ & $\delta$1$\uparrow$ & $\delta$2$\uparrow$  & $\delta$3$\uparrow$  \\
			\midrule
			MonoDepth2~\cite{godard2019digging} & \multirow{8}[1]{*}{SemanticKITTI} & 0.5259 & 7.113 & 14.43 & 1.0292 & 10.44 & 26.32 & 41.43  \\
			SynSin~\cite{synsin}  & &   0.3611 & 3.483 & 8.824 & 0.4290 & 52.61 & 74.56 & 86.50   \\
			PixelNeRF~\cite{pixelnerf}  & &   0.2364 & 2.080 & 6.449 & 0.3354 & 65.81 & 85.43 & 92.90  \\
			MINE~\cite{mine2021}  & & 0.2248 & 1.787 & 6.343 & 0.3283 & 65.87	& 85.52 & 93.30  \\
			VisionNerf~\cite{visionnerf} & & 0.2054 & 1.490 & 5.841 & 0.3073 & 69.11 & 88.28 & 94.37  \\
			SceneRF~\cite{cao2023scenerf} &  & \underline{0.1681}	& \underline{1.291} & \underline{5.781} & \underline{0.2851} & \underline{75.07} & \underline{89.09} & \underline{94.50}  \\
            \bf{SelfOcc (TPV)} & & \bf{0.1562} & \bf{1.153} & \bf{5.349} & \bf{0.2534} & \bf{78.92} & \bf{91.12} & \bf{95.92} \\
            \midrule
            SceneRF~\cite{cao2023scenerf} & \multirow{2}[1]{*}{nuScenes} & 0.7631 & 15.34 & 11.50 & 0.6240 & 40.26 & 64.86 & 77.02 \\
            \bf{SelfOcc (TPV)}& & \bf{0.4003} & \bf{6.723} & \bf{8.460} & \bf{0.4165} & \bf{64.50} & \bf{80.40} & \bf{88.17}\\
			\bottomrule
		\end{tabular}
	}
	\label{tab: novel depth}
\vspace{-2mm}
\end{table*}

\begin{table*}[t]
	\centering	
	\caption{
 \textbf{Self-supervised depth estimation on the nuScenes~\cite{caesar2020nuscenes} and KITTI-2015~\cite{kitti-2015} dataset.} 
 M: trained with monocular sequence. 
 MS: trained with both monocular sequence and stereo images. 
 SV: trained with surround-view images. 
 TO: temporal offline refinement. 
 We train SurroundDepth* with ground truth poses based on the official implementation.
	}
	\vspace{-3mm}
	\setlength{\tabcolsep}{0.008\linewidth}
	\resizebox{0.95\textwidth}{!}{
		\begin{tabular}{l|c|cc|ccccccc}
			\toprule
			Method & Dataset & Type & Scale Method & Abs Rel$\downarrow$ & Sq Rel$\downarrow$ & RMSE$\downarrow$ & RMSE log$\downarrow$ & $\delta$1$\uparrow$ & $\delta$2$\uparrow$  & $\delta$3$\uparrow$ \\
			\midrule
			Monodepth2~\cite{godard2019digging} & \multirow{5}[1]{*}{nuScenes} & SV & Median &  0.287  &   \underline{3.349}  &   \underline{7.184}  &   \underline{0.345}  &   0.641  &   0.845  &   \underline{0.925}
			\\
			FSM~\cite{guizilini2021full} &  & SV & Median &   0.299  &   -  &   -  &   -  &   -  &   -  &   -
			\\
			SurroundDepth~\cite{wei2023surrounddepth} & & SV  & SfM Pretrain  & \underline{0.280}  &   4.401  &  7.467  & 0.364 & 0.661  &  0.844  & 0.917
			\\
            SurroundDepth*~\cite{wei2023surrounddepth} & & SV   & Pose GT &   0.342  &   7.917  &   8.019  &   0.360  &   \underline{0.716}  &   \underline{0.858}  &   0.918
            \\
			\bf{SelfOcc (TPV)} & & SV   & Pose GT & \bf{0.215} & \bf{2.743}   & \bf{6.706}  & \bf{0.316}   & \bf{0.753}  & \bf{0.875} & \bf{0.932}
			\\
            \midrule
            R3D3~\cite{schmied2023r3d3} & nuScenes & TO & Extrinsics & 0.253 & 4.759 & 7.150 & - & 0.729 & - & - \\
            \midrule
            Johnston~\cite{johnston2020self} & \multirow{6}[1]{*}{KITTI-2015}  & M &  Median & 0.106 & 0.861 & 4.699 & 0.185 & \underline{0.889} & 0.962 & 0.982 \\
            FeatDepth(R50)~\cite{shu2020feature} &  & M &  Median & 0.104 & \underline{0.729} & 4.481 & \underline{0.179} & \textbf{0.893} & \underline{0.965} &  \underline{0.984} \\ 
            PackNet-SfM~\cite{guizilini20203d} &  & M & Median  & 0.107 & 0.802 & 4.538 & 0.186 & \underline{0.889} & 0.962 & 0.981\\ 
            R-MSFM6~\cite{zhou2021r} &  & M & Median  & 0.108 & 0.748 & \underline{4.470} & 0.185 & \underline{0.889} & 0.963 & 0.982\\ 
            DevNet~\cite{zhou2022devnet} &  & M & Median  & \textbf{0.100} & \textbf{0.699} & \textbf{4.412} & \textbf{0.174} & \textbf{0.893} & \textbf{0.966} & \textbf{0.985}  \\ 
            \bf{SelfOcc (TPV)} & & M  & Pose GT & \underline{0.103}  &       0.792  &       4.673  &       0.187  &       0.877  &       0.959  &       0.982 \\
            \midrule
            MonoDepth2(R50)~\cite{godard2019digging} & \multirow{6}[1]{*}{KITTI-2015} & MS & Extrinsics  & 0.106 & 0.818 & 4.75 & 0.196 & 0.874 & 0.957& 0.979  \\ 
            FeatDepth(R50)~\cite{shu2020feature} &  & MS & Extrinsics   & \underline{0.099} & \underline{0.697} & 4.427 & \underline{0.184} & \underline{0.889} & \underline{0.963} & \underline{0.982} \\ 
            R-MSFM6~\cite{zhou2021r} &  & MS & Extrinsics  & 0.108 & 0.753 & 4.469 & 0.185 & 0.888 & \underline{0.963} & \underline{0.982} \\ 
            DevNet~\cite{zhou2022devnet} &  & MS & Extrinsics  & \textbf{0.095} & \textbf{0.671} & \textbf{4.365} & \textbf{0.174} & \textbf{0.895} & \textbf{0.970}  & \textbf{0.988} \\ 
            BTS~\cite{wimbauer2023behind} & & MS  & Pose GT & 0.102 & 0.751 & \underline{4.407} & 0.188 &  0.882 &  0.961 &  \underline{0.982} \\
            \bf{SelfOcc (TPV)} & & MS  & Pose GT &       \underline{0.099}  &       0.711  &       4.586  &       0.186  &       0.880  &       0.960  &       \underline{0.982} \\
			\bottomrule
	\end{tabular}}
	\vspace{-5mm}
	\label{tab:depth estimation}
\end{table*}

\subsection{Task Descriptions}
\textbf{The 3D occupancy prediction task} aims to predict the dense occupancy (and semantic) states for a voxel grid, which is a fundamental task in autonomous driving due to its fine granularity and least abstraction of the real world.
We conduct this task on the Occ3D-nuScenes~\cite{tian2023occ3d} dataset, and use IoU and mIoU as geometric and semantic metrics, respectively.
In addition, we evaluate our method on the SemanticKITTI~\cite{behley2019semantickitti} dataset for the geometry-only 3D occupancy prediction task and report IoU, Precision and Recall.

\textbf{The novel depth synthesis task} synthesizes depth maps from novel viewpoints within a certain distance given the input frame~\cite{cao2023scenerf}.
This task is more reliable as a proxy for 3D reconstruction compared with novel view synthesis because the rendered color is an integration of predicted radiance at sampled locations, which introduces unnecessary degrees of freedom.
We conduct experiments on nuScenes~\cite{caesar2020nuscenes} and SemanticKITTI~\cite{behley2019semantickitti} and use Abs Rel, Sq Rel, RMSE, RMSE log and threshold accuracies ($\delta1$, $\delta2$, $\delta3$) as metrics.

\textbf{The depth estimation task} is a traditional task for 3D perception, which predicts the pixel-aligned depth map for a given frame.
We conduct surround-view depth estimation on nuScenes~\cite{caesar2020nuscenes} and monocular depth estiamtion on KITTI-2015~\cite{kitti-2015}.
The same metrics are used as in novel depth synthesis.
We detail the datasets in Section~\ref{app:datasets}.

\subsection{Implementation Details}
We use the identical network architecture for all three tasks, while the temporal supervision and the loss formulation may differ according to their respective objectives.
We adopt ResNet50~\cite{he2016resnet} pretrained on ImageNet~\cite{deng2009imagenet} as the 2D backbone and an feature pyramid network (FPN)~\cite{lin2017fpn} to generate multi-scale image features.
The 3D encoder $\mathcal{F}$ is the same as in BEVFormer~\cite{li2022bevformer} or TPVFormer~\cite{huang2023tri} depending on the 3D representation we use.
We further employ a two-layer MLP as the decoder network $\mathcal{D}$ to generate the SDF field, color and semantic logits (optional). 

We activate temporal supervision for 3D occupancy prediction and novel view synthesis which both involve scene reconstruction from multiple viewpoints, while we use only the current frame as supervision for depth estimation since it predicts the geometry only for the input view.
As for loss formulation, we use $L_{dep}$, $L_{E}$ and an edge loss $L_{edg}$~\cite{godard2017unsupervised} for depth estimation since RGB supervision from the input view is meaningless, while $L_H$ and $L_s$ focus on improving the overall geometry, thus helpless for depth estimation.
For novel depth synthesis, we leverage $L_{dep}$, $L_{rgb}$ and $L_{E}$.
And for 3D occupancy prediction, we further activate $L_H$ and $L_s$ on the basis of novel depth synthesis to impose smoothness and sparsity constraints on the predicted occupancy grid.
For semantic prediction, we use the off-the-shelf open-vocabulary segmentation network OpenSeeD~\cite{openseed} to generate pseudo segmentation labels for nuScenes~\cite{caesar2020nuscenes}.
We use the AdamW~\cite{loshchilov2017adamw} optimizer with an initial learning rate of 1e-4 which decays to zero following the cosine schedule.
We train our models for 12 epochs on the nuScenes~\cite{caesar2020nuscenes} dataset and 24 epochs on the SemanticKITTI~\cite{behley2019semantickitti} and KITTI-2015~\cite{kitti-2015} datasets.

\begin{table}[t]
        \footnotesize	
        \centering
	\caption{\textbf{Architecture ablation on the nuScenes dataset.} $L_{mvs}$: the proposed $L_{mvs}$ ($\checkmark$) or $L_{rpj}$ for depth optimization. SDF: SDF field ($\checkmark$) or density field for volume rendering.
	}
	\vspace{-3mm}
	  \setlength{\tabcolsep}{0.01\linewidth}
		\begin{tabular}{cc|cc|cc|cc}
			\toprule
			\multirow{2}[1]{*}{$L_{mvs}$} & \multirow{2}[1]{*}{SDF} & \multicolumn{2}{c|}{Occ.} & \multicolumn{2}{c|}{Novel Depth} & \multicolumn{2}{c}{Depth} \\
            & & IoU & mIoU & Abs Rel$\downarrow$ & RMSE$\downarrow$ & Abs Rel$\downarrow$ & RMSE$\downarrow$ \\
			\midrule
             & \checkmark & 18.90 & 2.81 & 0.8101 & 12.408 & 0.655 & 13.242
			\\
            \checkmark &  & 36.68 & 6.10 & \bf{0.4007} & \bf{8.149}  & 0.318 & 8.524
            \\
            \checkmark & \checkmark & \bf{38.43} & \bf{6.94} & 0.4376  & 8.734  & \bf{0.312} & \bf{7.544} 
			\\
			\bottomrule
    	\end{tabular}
	\vspace{-3mm}
	\label{tab: arch ablation}
\end{table}

\begin{table}[t]
        \footnotesize
        \centering
	\caption{\textbf{Temporal supervision ablation on nuScenes~\cite{caesar2020nuscenes}.}
 We do not use semantic supervision for occupancy prediction here.
	}
	\vspace{-3mm}
	  \setlength{\tabcolsep}{0.008\linewidth}
		\begin{tabular}{ccc|ccc|cc|cc}
			\toprule
			\multicolumn{3}{c|}{Supervision Ratio} & \multicolumn{3}{c|}{Occ.} & \multicolumn{2}{c|}{Novel Depth} & \multicolumn{2}{c}{Depth} \\
            Curr. & Prev. & Next & IoU & Prec. & Rec. & Abs Rel & RMSE & Abs Rel & RMSE \\
			\midrule
            1 & 0 & 0 & 39.67 & 66.96 & 49.33 & 0.471 & 8.852 & \bf{0.312} & \bf{7.544}
			\\
            2 & 1 & 1 & 40.50 & 67.51 & 50.31 & 0.446 & 8.827 & 0.365 & 8.046
            \\
            1 & 1 & 1 & 40.47 & 63.98 & 52.41 & \bf{0.438} & 8.734 & 0.363 & 7.975
			\\
            1 & 2 & 2 & 40.43 & 63.35 & \bf{52.78} & 0.442 & 8.781 & 0.366 & 7.998
			\\
            0 & 1 & 1 & \bf{42.48} & \bf{70.08} & 51.89 & 0.462 & \bf{8.732} & 0.392 & 8.075 
            \\
			\bottomrule
    	\end{tabular}
	\vspace{-3mm}
	\label{tab: ablation temporal}
\end{table}

\begin{table}[t]
        \footnotesize
        \centering
	\caption{\textbf{Loss and regularization ablation} on the SemanticKITTI~\cite{behley2019semantickitti} (occupancy prediction and novel depth synthesis) and KITTI-2015~\cite{kitti-2015} (depth estimation) datasets. 
	}
	\vspace{-3mm}
	  \setlength{\tabcolsep}{0.0055\linewidth}
		\begin{tabular}{cccc|ccc|cc|cc}
			\toprule
			\multirow{2}[1]{*}{$L_{dep}$} & \multirow{2}[1]{*}{$L_{rgb}$} & \multirow{2}[1]{*}{$L_H$} & \multirow{2}[1]{*}{$L_s$} & \multicolumn{3}{c|}{Occ.} & \multicolumn{2}{c|}{Novel Depth} & \multicolumn{2}{c}{Depth} \\
            & & & & IoU & Prec. & Rec. & Abs Rel & RMSE & Abs Rel & RMSE \\
			\midrule
             & \checkmark & & & 9.81 & 9.85 & \bf{96.18} & 0.3852 & 11.67 & - & -
            \\
            \checkmark & & & & 23.51 & \bf{41.01} & 35.51 & 0.1579 & \bf{5.348} & 0.102 & \bf{4.676}
            \\
            \checkmark & \checkmark & & & \bf{24.07} & 37.78 & 39.88 & \bf{0.1573} & 5.352 & \bf{0.100} & 4.773
            \\
            \checkmark & \checkmark & \checkmark & & 12.27 & 13.02 & 68.14 & 0.1722 & 5.748 & 0.107 & 5.091
            \\
            \checkmark & \checkmark & \checkmark & \checkmark & 22.16 & 33.54 & 39.50 & 0.1694  & 5.601 & 0.113 & 5.078
			\\
			\bottomrule
    	\end{tabular}
	\vspace{-5mm}
	\label{tab:ablation loss}
\end{table}

\begin{figure}[!t]
    \centering
    \includegraphics[width=\linewidth]{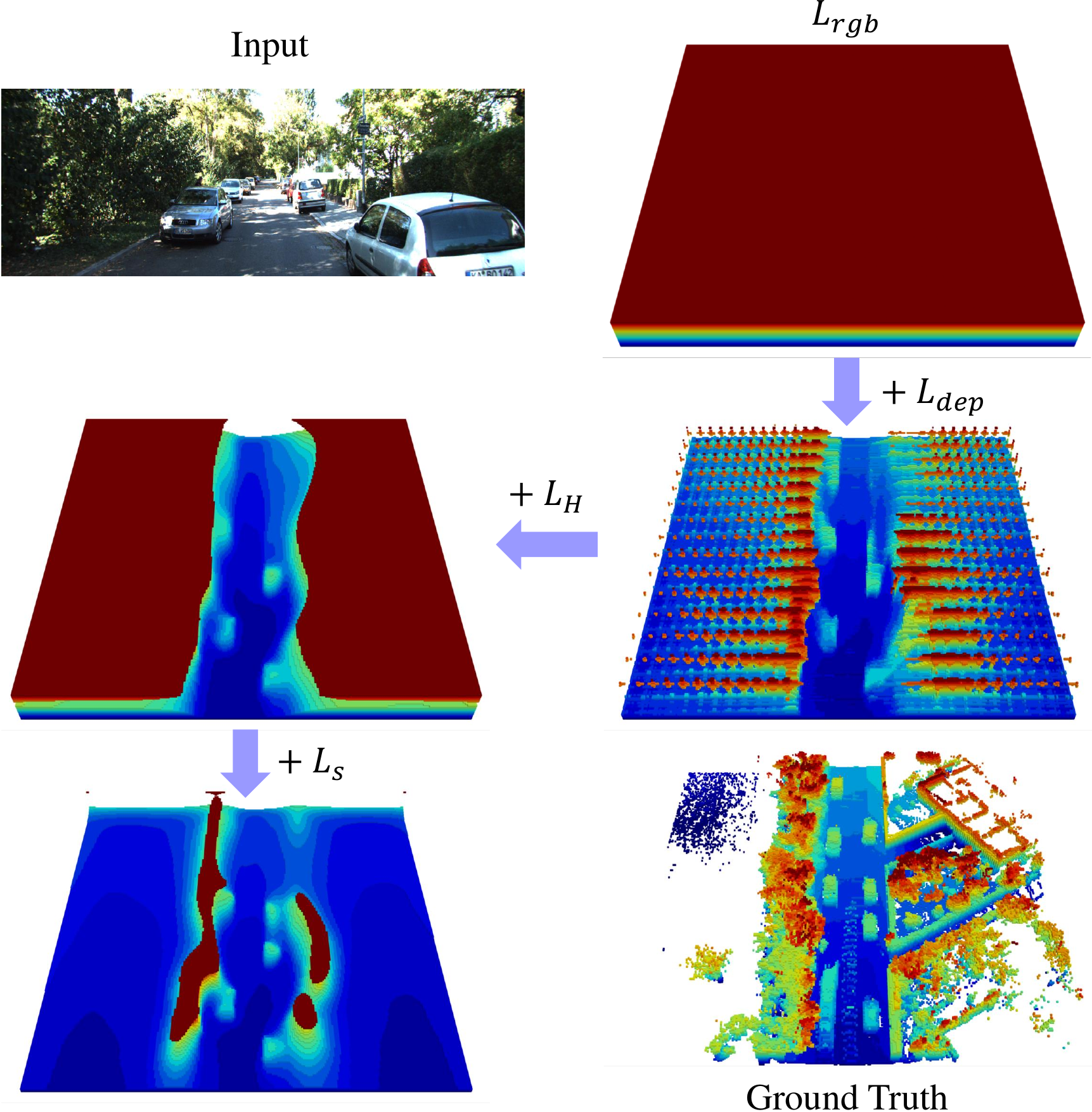}
	\vspace{-6mm}
    \caption{
    Visualization of the effect of different loss functions.
    The occupancy quality gradually improves with $L_{dep}$, $L_H$ and $L_s$.
    }
    \label{fig: vis loss}
	\vspace{-5mm}
\end{figure}

\begin{figure*}
    \centering
    \includegraphics[width=\linewidth]{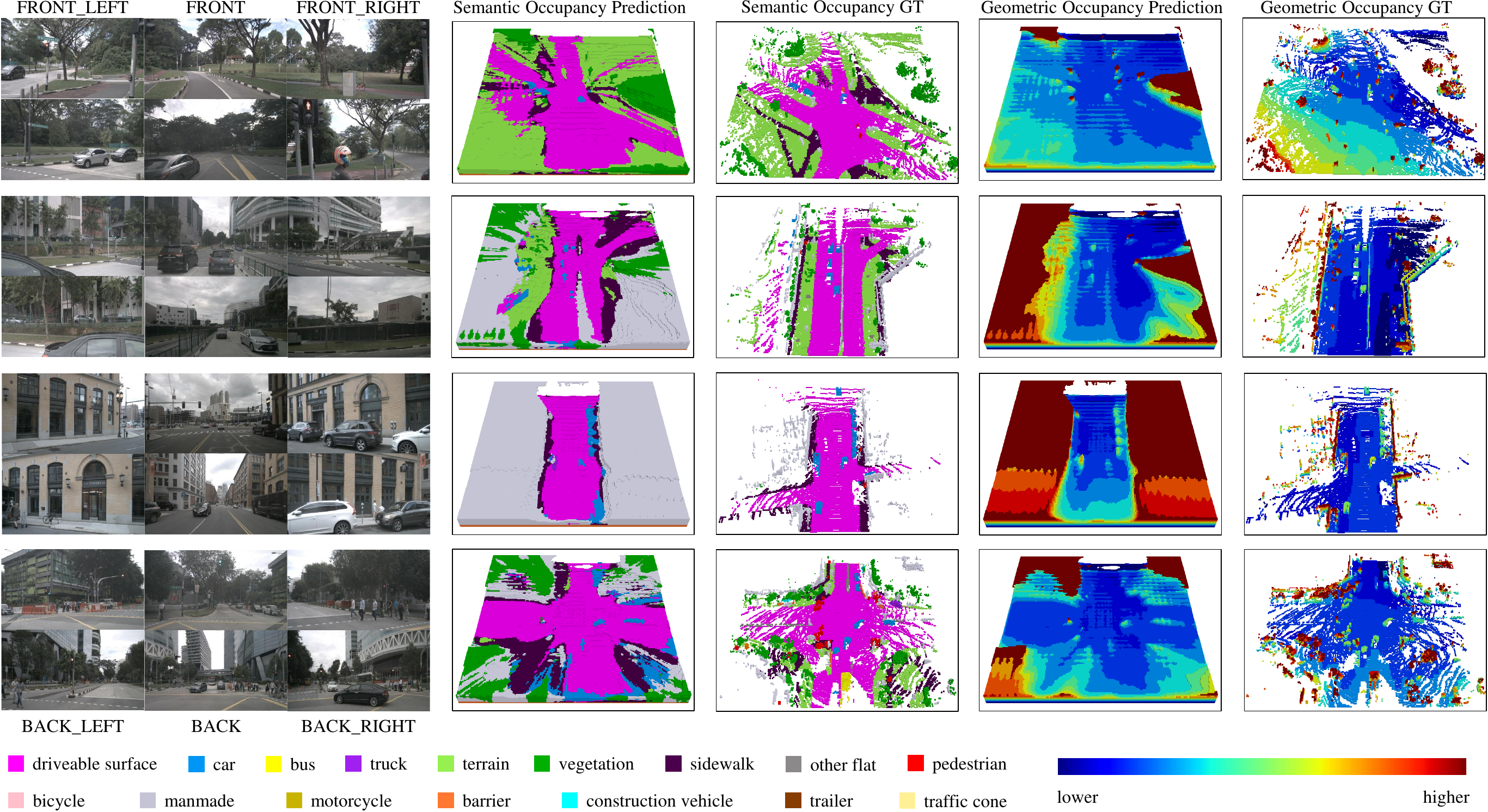}
	\vspace{-6mm}
    \caption{\textbf{Visualization results for 3D occupancy prediction on the Occ3D-nuScenes~\cite{tian2023occ3d} dataset.} 
    Our method achieves comparable visualization quality with ground truth for both semantic and geometric occupancy prediction tasks.
    }
    \label{fig: vis nuscenes occ}
	\vspace{-5mm}
\end{figure*}

\subsection{Results and Analysis}
\textbf{3D Occupancy Prediction.}
We report the results for the 3D occupancy prediction task in Table~\ref{table:occ3d-nus} and \ref{tab:semantickitti occ}.
On the Occ3D-nuScenes~\cite{tian2023occ3d} dataset, our methods achieve comparable IoU and reasonable mIoU without any form of 3D supervision compared with supervised approaches.
The IoU metric (45.01\%) indicates that SelfOcc learns meaningful geometry given only video sequences.
We notice that SelfOcc is better at predicting background classes than foreground classes, which might be related to the semantic misalignment of the open-vocabulary 2D segmentation network we use.
Nonetheless, our method still outperforms MonoScene~\cite{cao2022monoscene} supervised by ground truth in mIoU and LiDAR-supervised TPVFormer~\cite{huang2023tri} in IoU.

On the SemanticKITTI~\cite{behley2019semantickitti} dataset, we use the results reported in SceneRF~\cite{cao2023scenerf}, where baselines are grouped into 3D, depth and image supervised categories, and the superscription $^{rgb}$ indicates the adaptation from 3D input to image input.
From Table~\ref{tab:semantickitti occ}, SelfOcc sets the new state-of-the-art for vision-based depth-supervised and self-supervised 3D occupancy prediction tasks, which outperforms SceneRF~\cite{cao2023scenerf} by 58.7\% in IoU, with much higher Precision and slightly lower Recall.
Visualizations for 3D occupancy predictions are shown in Figure~\ref{fig: vis nuscenes occ}. 

\textbf{Novel Depth Synthesis.}
We report the results for novel depth synthesis in Table~\ref{tab: novel depth}, which averages depth metrics over all frames no further than 10 meters away from the input frame.
Following SceneRF~\cite{cao2023scenerf}, we compare against methods based on monocular depth estimation~\cite{godard2019digging}, generalizable NeRF~\cite{pixelnerf,mine2021,visionnerf} and 3D-aware GAN~\cite{synsin}.
Our method outperforms previous state-of-the-art SceneRF~\cite{cao2023scenerf} on all metrics on the SemanticKITTI~\cite{behley2019semantickitti} dataset.
In addition, we modify SceneRF for the nuScenes~\cite{caesar2020nuscenes} dataset by randomly selecting one camera every iteration during training, and directly report the depth metric for the input frame considering the complexity of volume rendering conditioned on image features from surround cameras.
SelfOcc still achieves better results than SceneRF, which confirms its superiority in 3D reconstruction from surround views.
We provide novel depth visualizations in Section~\ref{app:novel_depth}.

\textbf{Depth Estimation.}
We report the results for depth estimation in Table~\ref{tab:depth estimation}.
Our method achieves state-of-the-art performance on nuScenes for self-supervised surround-view depth estimation. 
We also train SurroundDepth~\cite{wei2023surrounddepth} using the official implementation with ground-truth poses to eliminate their influence, which indicates that ground-truth poses do not necessarily boost performance compared with median-scaling applied for scale-ambiguous methods.
Additionally, SelfOcc performs on par with the state-of-the-art for self-supervised monocular depth estimation on KITTI-2015~\cite{kitti-2015}.
Visualizations can be found in Section~\ref{app:depth}.

\subsection{Ablation Study}
We conduct comprehensive ablation studies on architectural choices, temporal supervision, and loss formulations to validate the effectiveness of every component of our method.
All experiments are trained for half of the original epochs.

\textbf{Architecture.}
We investigate the effect of $L_{mvs}$ and the SDF field in Table~\ref{tab: arch ablation}.
MVS-embedded depth optimization brings consistent improvement for all three tasks.
While the SDF field outperforms the density field in occupancy prediction and depth estimation, it achieves worse results in novel depth synthesis.
We think it is because the SDF field holds an inherent smoothness prior and is thus more difficult to optimize for novel depth synthesis.

\textbf{Temporal Supervision.}
We report the results of different temporal supervision settings in Table~\ref{tab: ablation temporal}.
It is shown that increasing temporal supervision benefits occupancy prediction, but deteriorates depth estimation performance.
In addition, the best score for novel depth synthesis is achieved with a moderate level of temporal supervision (1:1:1).

\textbf{Loss Formulations.}
We ablate the effect of loss functions in Table~\ref{tab:ablation loss}.
Note that we replace $L_{rbg}$ with $L_{edg}$ for depth estimation task.
Models trained with depth loss $L_{dep}$ and color loss $L_{rgb}$ achieve best performance for depth-related tasks.
As shown in Figure~\ref{fig: vis loss}, we use the Hessian loss $L_H$ and sparsity loss $L_s$ to regularize the SDF field for better visualization quality, although models trained without them can also achieve high IoU during evaluation.
In fact, the sparsity loss makes the prediction more similar to the ground truth, but it may not improve the IoU under the evaluation protocol of using camera masks as in Occ3D~\cite{tian2023occ3d}.

\subsection{3D Occupancy Visualization}
We provide the visualization results for 3D occupancy prediction task on nuScenes~\cite{caesar2020nuscenes} in Figure~\ref{fig: vis nuscenes occ}.
Despite being trained with video sequences only, our model produces meaningful geometric and reasonable semantic predictions compared with ground truth.

\section{Conclusion}
In this paper, we have presented a self-supervised vision-based 3D occupancy prediction method (SelfOcc) for autonomous driving, which predicts meaningful geometry and semantics from video sequences only.
To facilitate feature extraction in the 3D space, we have proposed to condition NeRFs on 3D representations.
Furthermore, we have designed an MVS-embedded strategy to boost depth optimization for NeRFs.
SelfOcc has set new state-of-the-art for self-supervised vision-based occupancy prediction on SemanticKITTI and has predicted reasonable 3D occupancy for surrounding cameras on Occ3D for the first time.

\begin{figure*}[!t]
\centering
\includegraphics[width=\textwidth]{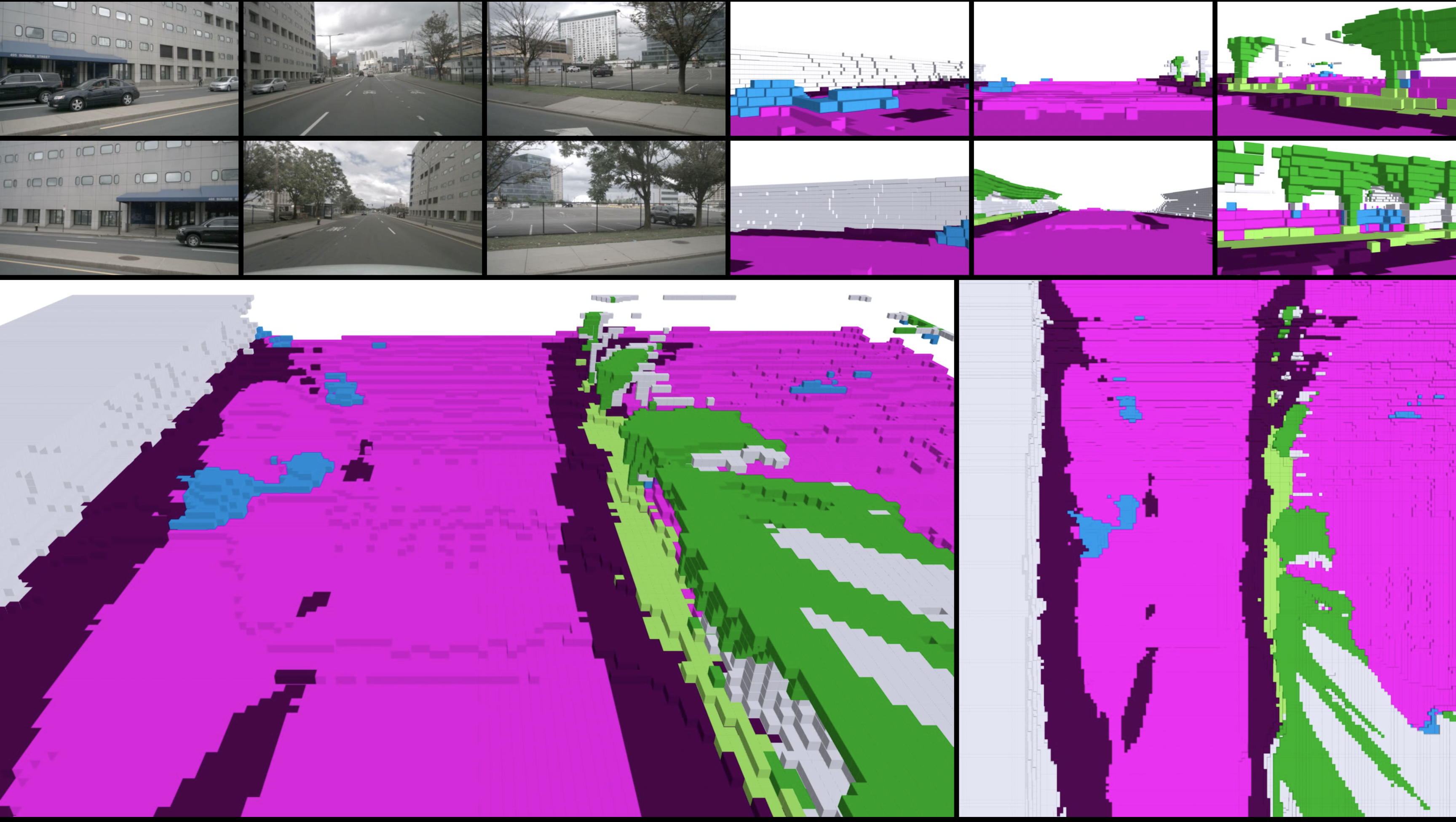}
\vspace{-6mm}
\caption{
\textbf{Visualizations of the proposed SelfOcc method for 3D semantic occupancy prediction on the nuScenes validation set.}
We show the six input surrounding images in the upper left and the predicted semantic occupancy from the corresponding views.
The lower parts demonstrate the predicted results in the global view (left) and bird's eye view (right). 
}
\label{teaser_supp}
\vspace{-5mm}
\end{figure*}

\appendix
\renewcommand{\tablename}{Tab}

\section{Dataset Details}
\label{app:datasets}

\textbf{The nuScenes~\cite{caesar2020nuscenes} dataset} consists of 1000 sequences of various driving scenes under different weather and lighting conditions, which are officially split into 700/150/150 sequences for training, validation and testing.
Each sequence lasts 20 seconds with LiDAR point cloud and RGB images collected by 6 surround cameras, and the keyframes are annotated at 2Hz.
In addition, the Occ3D-nuScenes~\cite{tian2023occ3d} dataset provides 3D semantic occupancy annotations with a resolution of 200x200x16 for 18 classes, covering an area of 80/80/6.4 meters around the ego car in the x/y/z-axis.

\textbf{The KITTI-2015~\cite{kitti-2015} dataset} holds stereo images from two forward-facing cameras and LiDAR point cloud.
Following \cite{godard2019digging}, we use the Eigen split~\cite{eigen2015predicting} and remove static images from the training set, which results in 39,810 monocular triplets for training and 4,424 for validation.

\textbf{The SemanticKITTI~\cite{behley2019semantickitti} dataset} is based on the odometry subset of the KITTI-2015~\cite{kitti-2015} dataset and provides voxelized lidar scans for 22 sequences with a resolution of 256x256x32.
Each voxel has a side length of 0.2m and is labeled with one of the 21 classes (19 semantic, 1 free and 1 unknown).
In our experiments, we only use the images from cam2 and follow the official split of the dataset, i.e. 10, 1 and 11 sequences for training, validation, and test.

\section{Additional Implementation Details}
\label{app: implementation}

\textbf{2D segmentor for semantic prediction.}
For 3D semantic occupancy prediction on nuScenes~\cite{caesar2020nuscenes}, we leverage the tiny version of the open-vocabulary 2D segmentor OpenSeeD~\cite{openseed} trained on COCO2017~\cite{lin2014coco} and Objects365v1~\cite{shao2019objects365} to directly predict semantic segmentation maps for supervision.
Note that although OpenSeeD ignores some classes due to rarity (e.g. construction vehicle) or semantic-ambiguity (e.g. others and other flat), we still consider all classes when calculating mIoU.

\textbf{Geometric settings.}
We use TPV~\cite{huang2023tri} or BEV~\cite{li2022bevformer} uniformly divided to represent a cuboid area, i.e. [80, 80, 6.4] meters around the ego car for nuScenes~\cite{caesar2020nuscenes} and [51.2, 51.2, 6.4] meters in front of the ego car for SemanticKITTI~\cite{behley2019semantickitti} and KITTI-2015~\cite{kitti-2015}.
The resolution for a single TPV/BEV grid cell is 0.4 meters for nuScenes and 0.2 meters for SemanticKITTI and KITTI-2015, respectively.
For depth prediction, we calculate metrics for depth values in the range of [0.1, 80] meters following \cite{godard2019digging,wei2023surrounddepth}.
And we evaluate depth prediction at 1:2 resolution against the raw image.

\textbf{Training settings.}
The resolution of input image is 384x800 for nuScenes, 370x1220 for SemanticKITTI following \cite{cao2023scenerf} and 320x1024 for KITTI-2015 following \cite{zhou2022devnet,godard2019digging}.
For the loss weights, we set $\lambda_c=\lambda_e=\lambda_H=0.1$, $\lambda_s = 0.001$ if present, and the weights for the edge $L_{edg}$ and the semantic $L_{sem}$ losses are 0.01 and 0.1, respectively, if applied.
We train our models on 8 RTX-3090 GPUs with 24GB memory.
Experiments on SemanticKITTI~\cite{behley2019semantickitti} and KITTI-2015~\cite{kitti-2015} take less than one day, while experiments on nuScenes~\cite{caesar2020nuscenes} finish within two days.

\section{Mathematical Derivation}
\label{app: math derivation}
In this section, we further discuss the advantage of our proposed MVS-embedded depth optimization over the traditional reprojection loss with mathematical derivations.

As in Eq.~(8), the reprojection loss can be formulated as
\begin{equation}\label{app eq: reproj}
    L_{rpj}(\mathbf{x}, \mathbf{I}_t, \mathbf{I}_s; \boldsymbol{\theta}) = \big\Vert\mathbf{I}_t(\mathbf{x}) - \mathbf{I}_s(\hat{\mathbf{x}}(\boldsymbol{\theta}))\big\Vert.
\end{equation}
Then we further expand $\mathbf{I}_s(\hat{\mathbf{x}}(\boldsymbol{\theta}))$ according to the definition of bilinear interpolation to get
\begin{equation}\label{app eq: reporj and bilinear}
    L_{rpj} = \big\Vert\mathbf{I}_t(\mathbf{x}) - \sum_{i, j \in \{0, 1\}} w_{ij}(\boldsymbol{\theta}) \mathbf{I}_s\big[\lfloor\hat{\mathbf{x}}+(i,j)\rfloor\big] \big\Vert, 
\end{equation}
where $\lfloor\cdot\rfloor$, $\lfloor\hat{\mathbf{x}}+(i,j)\rfloor$ and $\mathbf{I}_s[\cdot]$ denote the floor operation, the adjacent corner pixels of $\hat{\mathbf{x}}$ and the indexing operation, respectively.
In addition, $w_{ij}(\boldsymbol{\theta})$ is the normalized interpolation weight of the $ij$th adjacent corner pixel.
Note that once $\hat{\mathbf{x}}$ is calculated according to perspective transformation, $\mathbf{I}_s[\lfloor\hat{\mathbf{x}}+(i,j)\rfloor]$ is fixed and not differentiable with respect to $\boldsymbol{\theta}$.
Therefore, the receptive field of the optimization problem in (\ref{app eq: reporj and bilinear}) is limited to only four adjacent corner pixels involved in bilinear interpolation, which has an adverse effect on the efficiency and stability of depth learning.
Moreover, the summation operation in (\ref{app eq: reporj and bilinear}) is inside the norm bracket, which could lead to coupling of the adjacent corner pixels and local minima.
\begin{equation}\label{app eq: mvs depth}
    L_{mvs} = \sum_{m=1}^{M}w_m(\boldsymbol{\theta}) \Vert\mathbf{I}_t(\mathbf{x}) - \mathbf{I}_x(\pi(\mathbf{x}, d_m, \boldsymbol{\Pi}))\Vert.
\end{equation}
In contrast, our MVS-embedded depth optimization in (\ref{app eq: mvs depth}) moves the summation outside the dissimilarity metric, and effectively enlarges the receptive field by incorporating multiple depth candidates $d_m$ along the ray.

\begin{figure*}[t]
\centering
\includegraphics[width=\textwidth]{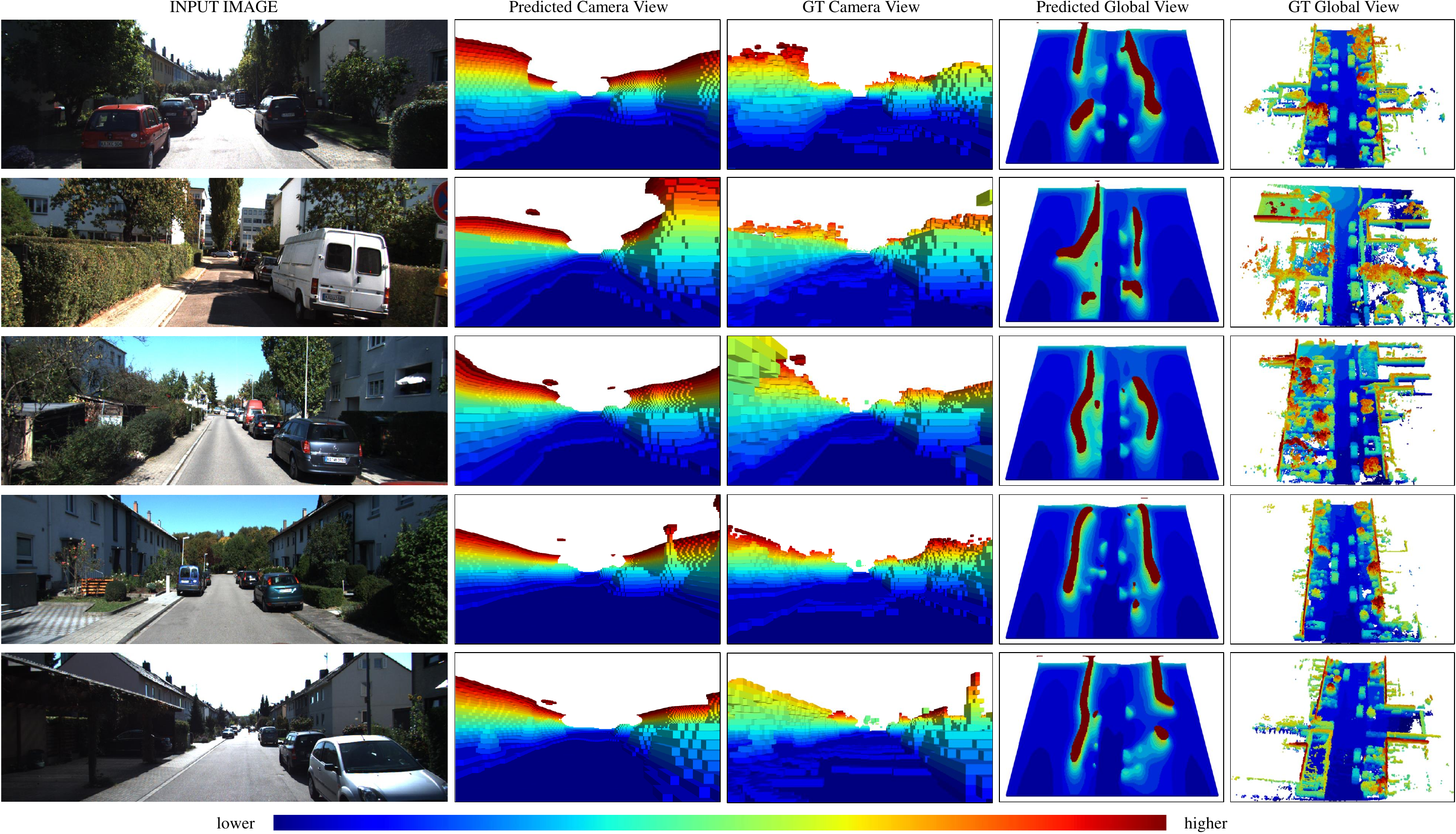}
\vspace{-6mm}
\caption{\textbf{Visualizations of 3D occupancy prediction on the SemanticKITTI~\cite{behley2019semantickitti} validation set.}
}
\label{app fig: kitti occ}
\vspace{-2mm}
\end{figure*}

\begin{figure*}[t]
\centering
\includegraphics[width=\textwidth]{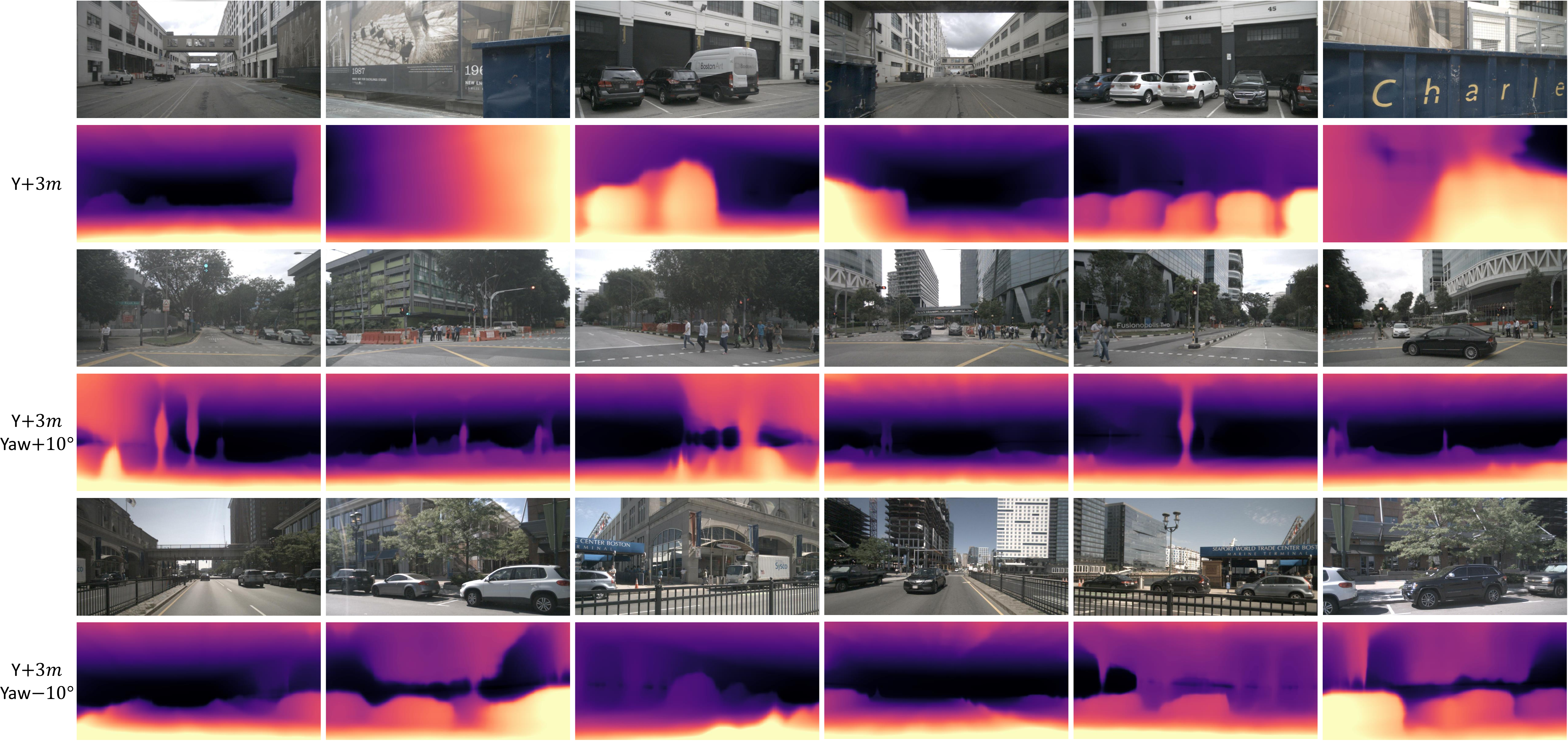}
\vspace{-6mm}
\caption{\textbf{Visualizations of novel depth synthesis on the nuScenes validation set.}
}
\label{novel_depth}
\vspace{-5mm}
\end{figure*}

\begin{figure*}[t]
\centering
\includegraphics[width=\textwidth]{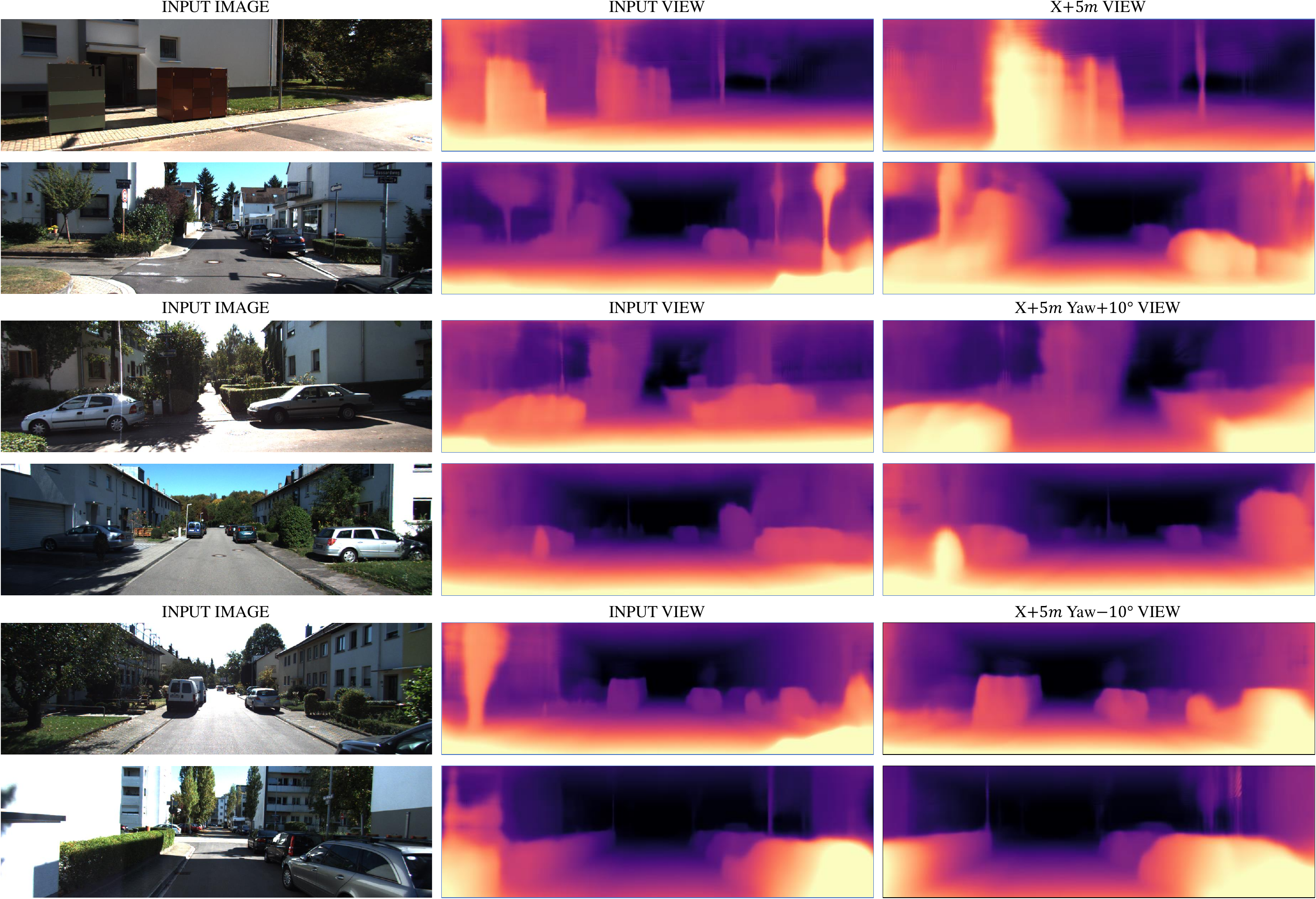}
\vspace{-6mm}
\caption{\textbf{Visualizations of novel depth synthesis on the SemanticKITTI validation set.}
}
\label{app fig: kitti novel depth}
\vspace{-2mm}
\end{figure*}

\begin{figure*}[t]
\centering
\includegraphics[width=\textwidth]{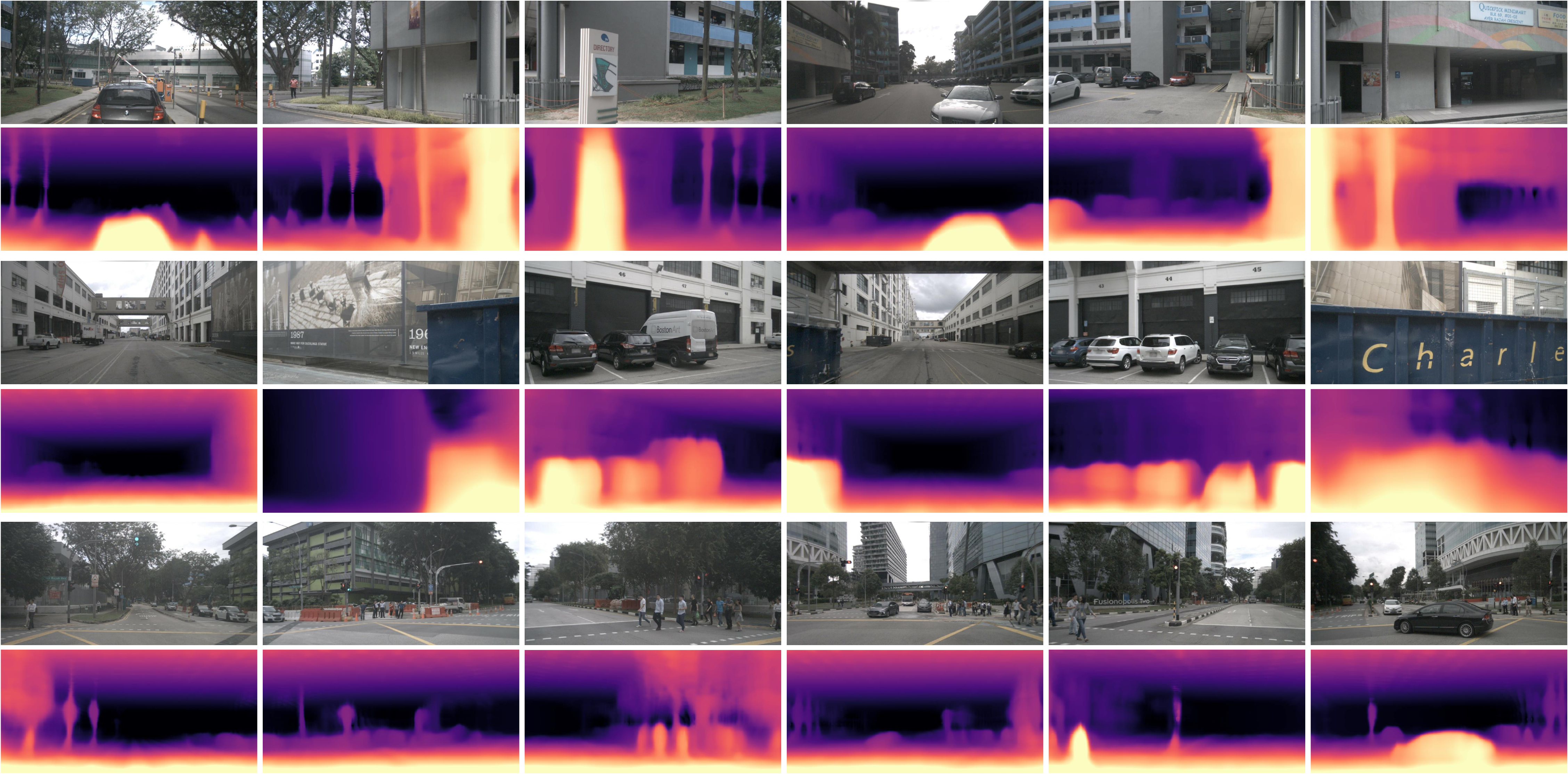}
\vspace{-6mm}
\caption{\textbf{Visualizations of surrounding depth prediction on the nuScenes validation set.}
}
\label{depth}
\vspace{-5mm}
\end{figure*}

\begin{figure*}[t]
\centering
\includegraphics[width=\textwidth]{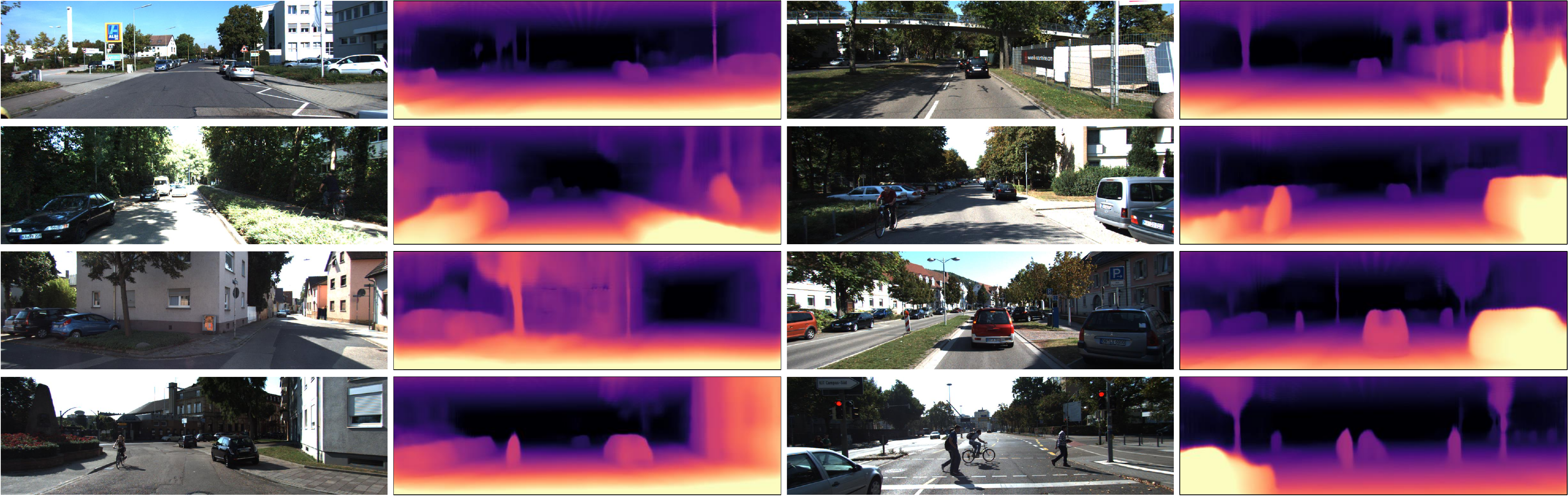}
\vspace{-5mm}
\caption{\textbf{Visualizations of depth estimation on the KITTI-2015~\cite{kitti-2015} test split.}
}
\label{app fig: kitti raw depth}
\vspace{-1mm}
\end{figure*}

\begin{figure*}[t]
\centering
\includegraphics[width=\textwidth]{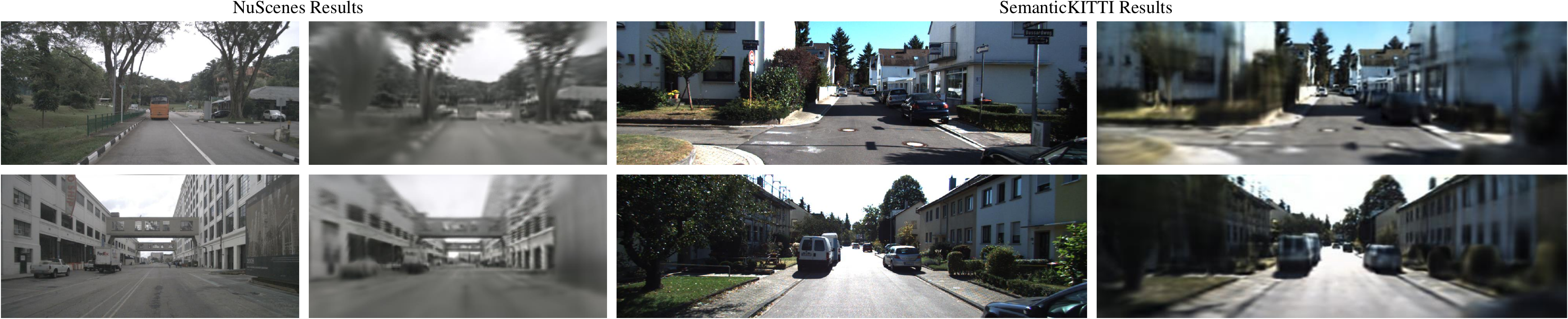}
\vspace{-6mm}
\caption{\textbf{Visualizations of novel view synthesis on nuScenes~\cite{caesar2020nuscenes} and SemanticKITTI~\cite{behley2019semantickitti}.}
We use the models for novel depth synthesis to synthesize novel views since these two tasks are similar.
SelfOcc suffers from the blurring effect.
}
\label{app fig: rgb synthesis}
\vspace{-5mm}
\end{figure*}

\section{Visualizations}

\subsection{3D Occupancy Prediction}
Figure~\ref{teaser_supp} shows a sampled image from the video demos~\footnote{\url{https://huang-yh.github.io/SelfOcc}.} for 3D geometric and semantic occupancy prediction on nuScenes~\cite{caesar2020nuscenes} validation set.
The demos show that SelfOcc can successfully infer semantic and geometric occupancy even for occluded areas.
Figure~\ref{app fig: kitti occ} shows the visualizations for 3D occupancy prediction on the SemanticKITTI~\cite{behley2019semantickitti} validation set, in which SelfOcc predicts accurate shapes and sizes of cars without any occupancy shadows.

\subsection{Novel Depth Synthesis}
\label{app:novel_depth}
 
Figure~\ref{novel_depth} and \ref{app fig: kitti novel depth} show the visualization results of novel depth synthesis on the nuScenes~\cite{caesar2020nuscenes} validation set and SemanticKITTI~\cite{behley2019semantickitti} validation set, respectively. 
Y$+3m$ (X$+5m$) means moving +3 (+5) meters along the y-axis (x-axis) of the LiDAR coordinate. 
Yaw$+10^{\circ}$/$-10^{\circ}$ means turning left/right for $10^{\circ}$.
SelfOcc trained with temporal supervision can predict 3D structures beyond the visible surface, thus generating high-quality novel depth views.

\subsection{Depth Estimation}
\label{app:depth}

Figure~\ref{depth} and \ref{app fig: kitti raw depth} shows the visualizations for depth estimation on the nuScenes~\cite{caesar2020nuscenes} validation set and KITTI-2015~\cite{kitti-2015} test split, respectively.
In addition to vehicles, our method successfully predicts sharp and accurate depth even for thin poles, moving pedestrians and cyclists.

\section{Limitations and Future Work}
Although we use color supervision during training to better exploit texture priors of RGB images, our model cannot synthesize high-quality novel views, suffering from the blurring effect as shown in Figure~\ref{app fig: rgb synthesis}, which is a long-standing problem in the field of generalizable NeRFs~\cite{pixelnerf,visionnerf,cao2023scenerf}.
In addition, although SelfOcc can predict accurate occupancy and depth for moving objects just as well as static ones, we do not include specific designs for motion.
We think the model might generalize the knowledge it learns from static elements to non-static ones.
Therefore, high-quality novel view synthesis and motion awareness could be potential focuses of future work.

{
    \small
    \bibliographystyle{ieeenat_fullname}
    \bibliography{main}
}

\end{document}